\documentclass[lettersize,journal]{IEEEtran}
\usepackage{amsmath,amsfonts}
\usepackage{algorithmic}
\usepackage{algorithm}
\usepackage{array}
\usepackage[caption=false,font=normalsize,labelfont=sf,textfont=sf]{subfig}
\usepackage{textcomp}
\usepackage{stfloats}
\usepackage{url}
\usepackage{verbatim}
\usepackage{graphicx}
\usepackage{cite}
\usepackage{xcolor}
\usepackage{booktabs}
\usepackage{enumitem}
\usepackage{orcidlink}

\graphicspath{{figs/}{figures/}{pictures/}{images/}{./}} 

\usepackage{tabu}                      
\usepackage{booktabs}                  
\usepackage{lipsum}                    
\usepackage{mwe}                       
\usepackage{float}
\usepackage{mathptmx}                  
\usepackage{amsmath,amsfonts,amssymb}
\newcommand{\norm}[1]{\left\lVert#1\right\rVert}

\usepackage{url}
\usepackage{comment}

\usepackage[capitalise,noabbrev,nameinlink]{cleveref}
\crefname{table}{Tab.}{Tabs.}
\Crefname{table}{Table}{Tables}
\crefname{section}{Sec.}{Secs.}
\Crefname{section}{Section}{Sections}
\crefname{figure}{Fig.}{Figs.}
\Crefname{figure}{Figure}{Figures}
\crefname{equation}{Eq.}{Eqs.}
\Crefname{equation}{Equation}{Equations}

\usepackage{tabularx}

\begin{document}
\definecolor{minor}{rgb}{0, 0.0, 0.0}

\title{MAPLE: Self-Supervised Learning-Enhanced Nonlinear Dimensionality Reduction \texorpdfstring{\\}{ } for Visual Analysis}

\author{
  Zeyang Huang,
  Takanori Fujiwara,
  Angelos Chatzimparmpas,
  Wandrille Duchemin, and
  Andreas Kerren
    \thanks{Zeyang Huang is with Link{\"o}ping University, Sweden. E-mail: zeyang.huang@liu.se.}
    \thanks{Takanori Fujiwara is with the University of Arizona, United States. E-mail: tfujiwara@arizona.edu.} 
    \thanks{Angelos Chatzimparmpas is with Utrecht University, Netherlands. E-mail: a.chatzimparmpas@uu.nl.}
    \thanks{Wandrille Duchemin is with University of Basel, Switzerland. E-mail: wandrille.duchemin@unibas.ch.}
    \thanks{Andreas Kerren is with Link{\"o}ping University and Linnaeus University, Sweden. E-mail: andreas.kerren@$\{$liu/lnu$\}$.se.}
}
\markboth{\textit{The final version of this paper is available at: \href{https://doi.org/10.1109/TVCG.2026.3694000}{10.1109/TVCG.2026.3694000}}}%
{Huang \MakeLowercase{\textit{et al.}}: MAPLE}

\maketitle

\begin{abstract}
We present a new nonlinear dimensionality reduction method, \textit{MAPLE}, that enhances UMAP by improving manifold modeling.
MAPLE employs a self-supervised learning approach to more \textit{efficiently} encode low-dimensional manifold geometry.
Central to this approach are \textit{maximum manifold capacity representations (MMCRs)}, which help untangle complex manifolds by compressing variances among locally similar data points while amplifying variance among dissimilar data points.
This design is particularly effective for high-dimensional data with substantial intra-cluster variance and curved manifold structures, such as biological or image data. 
Our qualitative and quantitative evaluations demonstrate that MAPLE can produce clearer visual cluster separations and finer subcluster resolution than UMAP while maintaining \textcolor{minor}{a tractable} computational cost.
\end{abstract}





\section{Introduction}
\newcommand{\OriginalData}{\mathbf{X}}
\newcommand{\Layout}{\mathbf{Y}}
\newcommand{\EmbedData}{\mathbf{Z}}
\newcommand{\NormalizedEmbedData}{\hat{\EmbedData}}
\newcommand{\Graph}{G}
\newcommand{\nPoints}{N}
\newcommand{\nDimensions}{D}
\newcommand{\nEmbedDimensions}{D'}
\newcommand{\nLayoutDimensions}{M}
\newcommand{\EncoderNetwork}{f_\theta}
\newcommand{\ProjectorNetwork}{g_\theta}
\newcommand{\Centroid}{\mathbf{C}}
\newcommand{\Members}{\mathcal{P}}
\newcommand{\Loss}{\mathcal{L}}
\newcommand{\EdgeWeight}{w_{ij}}

\IEEEPARstart{H}{igh-dimensional} datasets are rapidly growing across scientific domains, from single-cell transcriptomics to large-scale image collections.
Advances in genomics, machine learning, and related fields often produce measurements with tens of thousands of features, such as genes, pixels, or descriptors, that naturally place the data in very high dimensions.
As humans cannot directly see high-dimensional spaces, effective visualization is essential to analyze these datasets.
Dimensionality reduction (DR) methods make high-dimensional data visually explorable by projecting them into a low-dimensional space, typically 2D or 3D.
Effective DR results can facilitate data annotation, reveal rare populations, and lead to the generation of new hypotheses.

A wide range of DR methods has been proposed for visualization~\cite{espadoto2019toward,jeon2025unveiling}.
Linear DR methods, including principal component analysis (PCA)~\cite{hotelling1933analysis}, remain popular for their interpretability~\cite{cunningham2015linear,fujiwara2022ulca}, but are limited in capturing complex, curved data structures.
Nonlinear DR methods address this limitation by prioritizing local neighborhood relationships, allowing them to better represent data lying on complex manifolds~\cite{van2009dimensionality}.
Among these, uniform manifold approximation and projection (UMAP)~\cite{mcinnes_umap_2020, healy2024umap} has become a method of choice across many domains, alongside alternatives such as $t$-SNE~\cite{maaten2008visualizing}.
UMAP is valued for its scalability, graph-based formulation, and ability to capture nonlinear structure without strong parametric assumptions.
UMAP constructs a weighted $k$-nearest neighbor ($k$NN) graph in the input space to model an underlying manifold and then optimizes a low-dimensional layout so that the structure of the layout aligns with the graph according to a cross-entropy objective.
This procedure is computationally efficient, and its cross-entropy objective often yields layouts with clear cluster separation.

The quality of the resulting UMAP layout is strongly influenced by the quality of the weighted $k$NN graph. 
In practice, this graph may include connections that bring semantically distinct points close together in the low-dimensional layout, and vice versa.
We argue that a key limitation of UMAP lies not in the $k$NN algorithm itself---which can serve as an effective proxy for nonlinear structure---but in how the graph is constructed, specifically the choice of data representation and distance space used to define neighborhood relationships.
More specifically, standard $k$NN construction based on raw high-dimensional data and predefined distance metrics faces the following two issues.
First, as dimensionality increases, pairwise distances in high-dimensional space (e.g., Euclidean distances) tend to concentrate toward similar values~\cite{franccois2007concentration}, making $k$NN graphs increasingly sensitive to noise.
Second, many datasets lie on curved manifolds~\cite{kocabas2019self}, where points that appear as neighbors in the $k$NN graph may actually belong to distinct semantic regions~\cite{conwell2024large}.
In 2D or 3D layouts, these effects result in local clusters with spurious overlaps, thereby mixing groups that are semantically distinct.
Recent theoretical work has also analyzed these issues of UMAP~\cite{damrich_umap_2021, bohm2022attraction}, emphasizing its limitations in capturing subtle geometric differences.

To address the above fundamental challenges, 
we introduce a novel nonlinear DR method, \textit{MAPLE} (short for manifold-aware projection with learned edges).
A central idea in MAPLE is that explicitly learning neighborhood relationships can substantially improve the quality of low-dimensional layouts.
MAPLE refines neighborhood definitions by adapting multi-view self-supervised learning (MVSSL)~\cite{tsai2021selfsupervised, galvez2023role} to the context of nonlinear DR.
Unlike standard methods that treat a weighted $k$NN graph directly constructed from raw high-dimensional data as a fixed representation of the underlying manifold, MAPLE treats the initial neighbor relationships as weak, tentative self-supervised signals. 
We employ a learning phase to iteratively refine these signals.
Instead of directly operating in the input high-dimensional space, we employ encoder and projector networks~\cite{grill2020bootstrap,kalantidis2021tldr,zbontar_barlow_2021} to map data into an embedding space where the distance relationships are learned rather than fixed by a predefined distance metric.
A refined graph is then constructed based on these optimized distance relationships.
The core of this learning phase is the \textit{maximum manifold capacity representation (MMCR)} objective~\cite{yerxa2023mmcr}, which compresses variance among locally similar points (i.e., neighbors) while amplifying variance across dissimilar ones (i.e., non-neighbors). 
The outcome of this learning phase is a refined, weighted $k$NN graph that is subsequently laid out in a low-dimensional space using UMAP's cross-entropy optimization.

We evaluate MAPLE against multiple DR methods using a set of quality measures.
Additionally, we qualitatively evaluate MAPLE through visual analysis on image classification datasets.
We further demonstrate MAPLE's utility in case studies on single-cell transcriptomics in collaboration with a bioinformatics expert (a co-author of this work). 
Our results show that MAPLE reveals fine-grained, meaningful patterns that support deeper downstream analysis.

Our contributions are as follows:
\begin{itemize}[itemsep=2pt,topsep=2pt]
    \item MAPLE, a new nonlinear DR method that learns neighborhood structure through MVSSL;
    \item the adaptation of MMCRs to graph-based DR to untangle a complex underlying manifold through compression and diversification of local manifolds; and
    \item a set of quantitative metrics and qualitative evaluations using multiple datasets across domains, showing that MAPLE produces high-quality layouts.
\end{itemize}

\vspace{5pt}
\noindent\textbf{Terminology.}
MAPLE produces two types of low-dimensional representations. 
To avoid ambiguity, we use the following terminology throughout this paper:
\begin{description}[font=\normalfont,topsep=2pt,itemsep=2pt,leftmargin=18pt]
    \item[\textit{Embedding}:] an intermediate, relatively low-dimensional representation produced before the layout optimization.
    In MAPLE, embeddings are the outputs of the encoder--projector networks, and the weighted $k$NN graph is constructed and refined in this embedding space.
    \item[\textit{Layout}:] the final low-dimensional representation (typically 2D or 3D) obtained after the layout optimization phase of DR.
\end{description}

\vspace{5pt}\noindent
The remainder of this paper is organized as follows: \cref{sec:background} discusses background and related work. 
\cref{sec:maple} introduces MAPLE. 
\cref{sec:results} presents our experimental results, while \cref{sec:cases} presents case studies with single-cell datasets. 
\cref{sec:discussion} presents the limitations of our approach and opportunities for future work. 
In \cref{sec:conclusion}, we conclude our paper.

\section{Background and Related Work}
\label{sec:background}
DR aims to extract low-dimensional representations from high-dimensional data while preserving meaningful structure.
Nonlinear DR methods~\cite{van2009dimensionality} are designed to capture data patterns lying on nonlinear manifolds (e.g., curved surfaces). 
From the perspective of manifold modeling, nonlinear DR methods can be categorized by how they represent neighborhood structure. 
Graph-based modeling constructs a $k$NN graph or a similar structure to capture local proximities along a manifold. 
This category includes early spectral and geodesic methods, such as ISOMAP~\cite{tenenbaum2000global} and Laplacian eigenmaps~\cite{belkin2003laplacian}, as well as more recent methods, including LargeVis~\cite{tang2016visualizing}, $t$-SNE~\cite{maaten2008visualizing}, and UMAP~\cite{mcinnes_umap_2020}.
Diffusion-based modeling (e.g., spectral embeddings~\cite{meila2001spectral} and PHATE~\cite{moon_PHATE_2019}) simulates diffusion processes (e.g., random walks) on the data and uses transition probabilities as similarities to preserve manifold structure.
Kernel-based modeling~\cite{mika1998kernel, gisbrecht2015parametric} employs kernel functions to implicitly map data into higher-dimensional feature spaces, enabling the modeling of complex shapes.
Our new method, MAPLE, belongs to the graph-based family.
We specifically aim to advance graph-based modeling, using UMAP (a widely used state-of-the-art method) as our primary baseline.

\subsection{UMAP}
\label{sec:umap}

UMAP has been widely used in biology~\cite{becht2019dimensionality}, computational social science~\cite{garson2022factor}, and other data-intensive fields~\cite{espadoto2019toward}, due to its scalability, visually appealing layouts, and robust behavior across disciplines. 
The key idea of UMAP is to approximate the manifold structure of high-dimensional data by first constructing a weighted $k$NN graph, called a fuzzy graph, in the input space and then computing a low-dimensional layout that preserves the graph's structure.
McInnes et al.~\cite{mcinnes_umap_2020} describe these two phases as (1) \textbf{graph construction} and (2) \textbf{graph layout}.
MAPLE follows the same two-phase structure, while it refines the graph construction phase using MVSSL with MMCRs.
For clarity, we next provide a concise formal description of UMAP's two phases, which is necessary to highlight MAPLE's unique contributions.

\textbf{Phase 1: Graph construction.}
Let $\OriginalData = [\MakeLowercase{\OriginalData}_1 \cdots \MakeLowercase{\OriginalData}_\nPoints]^\top \in \mathbb{R}^{\nPoints \times \nDimensions}$ denote the original input data, where $\nPoints$ and $\nDimensions$ are the numbers of data points and dimensions, respectively.
In this phase, UMAP first constructs a $k$NN graph, $\Graph$, from $\OriginalData$ based on a chosen distance function $\mathrm{dist}(\cdot)$ (Euclidean distance by default).
UMAP then converts this graph into a fuzzy graph by assigning edge weights as follows:
\begin{equation}
\label{eq:umap_weight}
    \EdgeWeight = \exp({-\max(0,\, \mathrm{dist}(\MakeLowercase{\OriginalData}_{i}, \MakeLowercase{\OriginalData}_{j})-\rho_{i})} / \sigma_{i})
\end{equation}
where $\rho_i$ and $\sigma_i$ are local connectivity parameters automatically determined by UMAP to distribute data points uniformly over the manifold.
The weight, $\EdgeWeight$, reflects the strength of neighborhood relationships between points $i$ and $j$ (precisely, the membership strength of a 1-simplex between them~\cite{mcinnes_umap_2020}). 

\textbf{Phase 2: Graph layout.} \quad
Let $\Layout = [\MakeLowercase{\Layout}_1 \cdots \MakeLowercase{\Layout}_\nPoints]^\top \in \mathbb{R}^{\nPoints \times \nLayoutDimensions}$ denote the layout positions in an $\nLayoutDimensions$-dimensional space (usually, $\nLayoutDimensions = 2$ or $3$).
UMAP initializes $\Layout$ using a certain strategy, such as random initialization, PCA of $\OriginalData$, or spectral embedding of $\OriginalData$---with spectral embedding as the default.
Next, the edge weight between two points in the layout space is defined as: $v_{ij} = (1 + a \norm{\MakeLowercase{\Layout}_i - \MakeLowercase{\Layout}_j}^{2b})^{-1}$, where $a$ and $b$ are user-specified hyperparameters controlling how tightly neighbors are placed together.
In practice, $a$ and $b$ are derived from a single hyperparameter, \texttt{min\_dist}.
UMAP then optimizes $\Layout$ to minimize the cross-entropy loss between $\EdgeWeight$ and $v_{ij}$ through: 
\begin{equation}
\arg\min_{\Layout}
\sum_{i \neq j}
\Big[
\EdgeWeight \log\!\left(\frac{\EdgeWeight}{v_{ij}}\right)
+ (1 - \EdgeWeight)
\log\!\left(\frac{1 - \EdgeWeight}{1 - v_{ij}}\right)
\Big].
\end{equation}
To perform this optimization, UMAP first symmetrizes the fuzzy graph as $\EdgeWeight \gets \EdgeWeight + w_{ji} - \EdgeWeight w_{ji}$ and then applies negative sampling with stochastic gradient descent.

\textbf{Limitations of the graph construction phase.}
The cross-entropy loss encourages the similarity in the layout ($v_{ij}$) to be high when the high-dimensional weight ($\EdgeWeight$) is large (corresponding to an attractive force), and low when $\EdgeWeight$ is small (a repulsive force)~\cite{mcinnes_umap_2020, bohm2022attraction}.
Because of this formulation, the quality of the constructed graph directly determines the quality of the resulting layout. 
Damrich and Hamprecht~\cite{damrich_umap_2021} further showed that UMAP's stochastic negative sampling strongly down-weights the repulsive force, which increases the layout's sensitivity to changes in neighborhood connections.
In other words, the initial construction of the unweighted $k$NN graph is a critical factor in determining the quality of the final layout.

As described in \cref{eq:umap_weight}, UMAP assumes that data points are \textit{uniformly distributed} on the manifold under some Riemannian metric. 
Under this assumption, any ball of fixed volume should contain approximately the same number of points, regardless of its position on the manifold.
This motivates the use of fixed-size $k$NNs to represent local relationships.
In practice, however, real data are rarely uniform: they often vary in density or lie on curved manifolds.
As a result, $k$NNs constructed with a conventional metric (e.g., Euclidean distance) and the connectivity parameters, $\rho_i$ and $\sigma_i$, derived from them can misrepresent geodesic proximity. 
Since UMAP's cross-entropy objective optimizes only over the weighted $k$NN graph, any mismatch introduced during graph construction directly propagates into the final layout.
This observation highlights a research gap: improving neighborhood construction is essential for achieving low-dimensional layouts that faithfully reflect the underlying manifold geometry.

\subsection{Algorithmic Improvements of UMAP} 
Several works have refined UMAP's pipeline, either by modifying its loss formulation to better capture data characteristics or by guiding its hyperparameter selection.
For example, Narayan et al.~\cite{narayan_assessing_2021} showed that UMAP tends to overly expand dense regions while overly contracting sparse regions. 
To mitigate this issue, they introduced DensMAP, which augments UMAP's cross-entropy with a density-preservation term so that the layout better reflects underlying data density.
Similarly, Islam and Fleischer~\cite{islam2025shape} analyzed UMAP's attraction--repulsion force dynamics and modified these forces to improve cluster consistency under random initialization.
Other works have focused on guiding hyperparameter choices. 
For instance, Xia et al.~\cite{xia2024statistical} introduced scDEED, a statistical framework that evaluates layout reliability by comparing local neighborhoods before and after the graph layout phase, and then tunes hyperparameters such as the number of neighbors, $k$, to improve trustworthiness. 
In general, these refinements improve UMAP's layout behavior, but unlike MAPLE, they do not address the critical limitation rooted in the graph construction.

\subsection{Self-supervised Learning and Integration into DR} \label{sec:background_ssl} 

Self-supervised learning (SSL) has become a dominant approach for learning representations from unlabeled data~\cite{liu2021self}.
Its central idea is to let the data provide its own training signals.
A large family of SSL methods is known as MVSSL.
MVSSL's key concept is to create multiple \textit{views} of the same data that capture it from different perspectives.
These views can be generated through data augmentations, such as cropping, masking, and rotating images, or they can come from different modalities, such as paired audio and video.
A machine learning model (e.g., a neural network) is then trained to produce robust embeddings that remain consistent across different views of the same data~\cite{galvez2023role}.
Depending on how this alignment is achieved, MVSSL methods can be categorized~\cite{schaeffer_towards_2024} into contrastive methods (e.g., SimCLR~\cite{chen2020simple}), clustering-based methods (e.g., DeepCluster~\cite{caron2018deep}), distillation-based methods (e.g., BYOL~\cite{grill2020bootstrap}), and redundancy-reduction methods (e.g., Barlow Twins~\cite{zbontar_barlow_2021}).

Recently, MMCR-based methods have emerged as a new family within MVSSL.
These methods are inspired by manifold capacity theory~\cite{chung2018classification}, which studies the capacity of a manifold to separate a random set of regions.  
Intuitively, this capacity reflects the linear separability of local manifolds that compose the global manifold.
MMCRs were first introduced by Yerxa et al.~\cite{yerxa2023mmcr}. 
MMCRs employ the nuclear norm~\cite{peng2016connections} as a tractable surrogate for manifold capacity.
Since the nuclear norm is defined as the sum of singular values of the data matrix, it provides an estimate of the effective rank of the corresponding manifold.
Minimizing the nuclear norm compresses individual local manifolds into lower-rank structures, while maximizing the nuclear norm of manifold centroids promotes their separation.
This geometric formulation is distinctive to MMCRs, compared to the other MVSSL families, as it directly ties embedding quality to manifold capacity rather than pairwise alignment~\cite{schaeffer_towards_2024}.

To the best of our knowledge, MAPLE is the first method that adopts MMCRs into graph-based modeling for DR.
Below, we discuss DR methods that adopt other SSL families.

\textbf{Integration of SSL into DR.} \quad
Several works have integrated SSL into DR to improve layout quality. 
These works share critiques of the graph construction phase, including unreliable distances in high-dimensional space~\cite{bohm2022unsupervised}, semantically unstable neighborhoods~\cite{hu2023your}, and distortion caused by curvature or density differences~\cite{kalantidis2021tldr, feng2024ncldr}.
SSL objectives can help learn more semantically meaningful results.
For example, $t$-SimCNE~\cite{bohm2022unsupervised} combines the contrastive learning method, SimCLR, with a $t$-SNE-like loss to improve similarity estimation of data points.
However, as it relies on SimCLR, $t$-SimCNE is only applicable to image datasets.
TLDR~\cite{kalantidis2021tldr} employs a single nearest neighbor as the self-supervised signal and Barlow Twins~\cite{zbontar_barlow_2021} as the SSL loss.
While effective for data compression and preprocessing, TLDR is not designed for visualization purposes. 
Moreover, its reliance on a fixed $1$-nearest neighbor graph prevents it from addressing misaligned neighborhood issues.
Similarly, NCLDR~\cite{feng2024ncldr} also uses nearest neighbors as signals, but its architecture is tailored to images.

In summary, existing SSL-integrated DR methods are either restricted to specific domains (e.g., image datasets) or designed for preprocessing rather than visualization. 
They do not provide a general solution for handling misaligned neighborhoods, diverse datasets, or visualization of high-dimensional data.

\section{Manifold-aware Projection \texorpdfstring{\\}{ }with Learned Edges}
\label{sec:maple}
In this section, we first outline UMAP's critical limitations, which motivate our work.
We then introduce \textit{MAPLE} (manifold-aware projection with learned edges).\footnote{The source code and supplementary materials can be found at \url{https://github.com/zeyh/MAPLE}.}

\begin{figure}[tb]
  \centering
  \includegraphics[width=1\columnwidth]{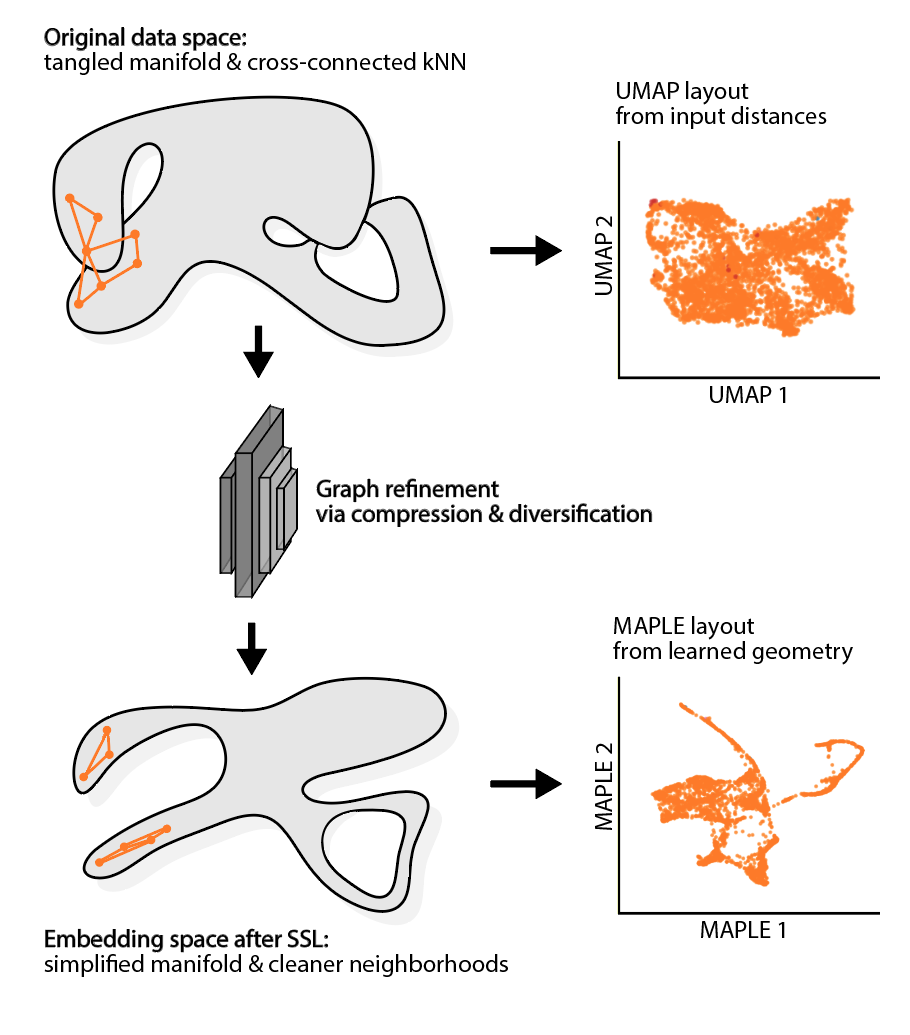}
  \caption{Conceptual overview of MAPLE. Self-supervised neighborhood refinement turns a cross-connected, entangled $k$NN graph into a simplified geometry that supports clearer graph layouts.}
  \label{fig:intro}
\end{figure}

\begin{figure*}[ht!]
\centering
    \includegraphics[width=0.85\linewidth]{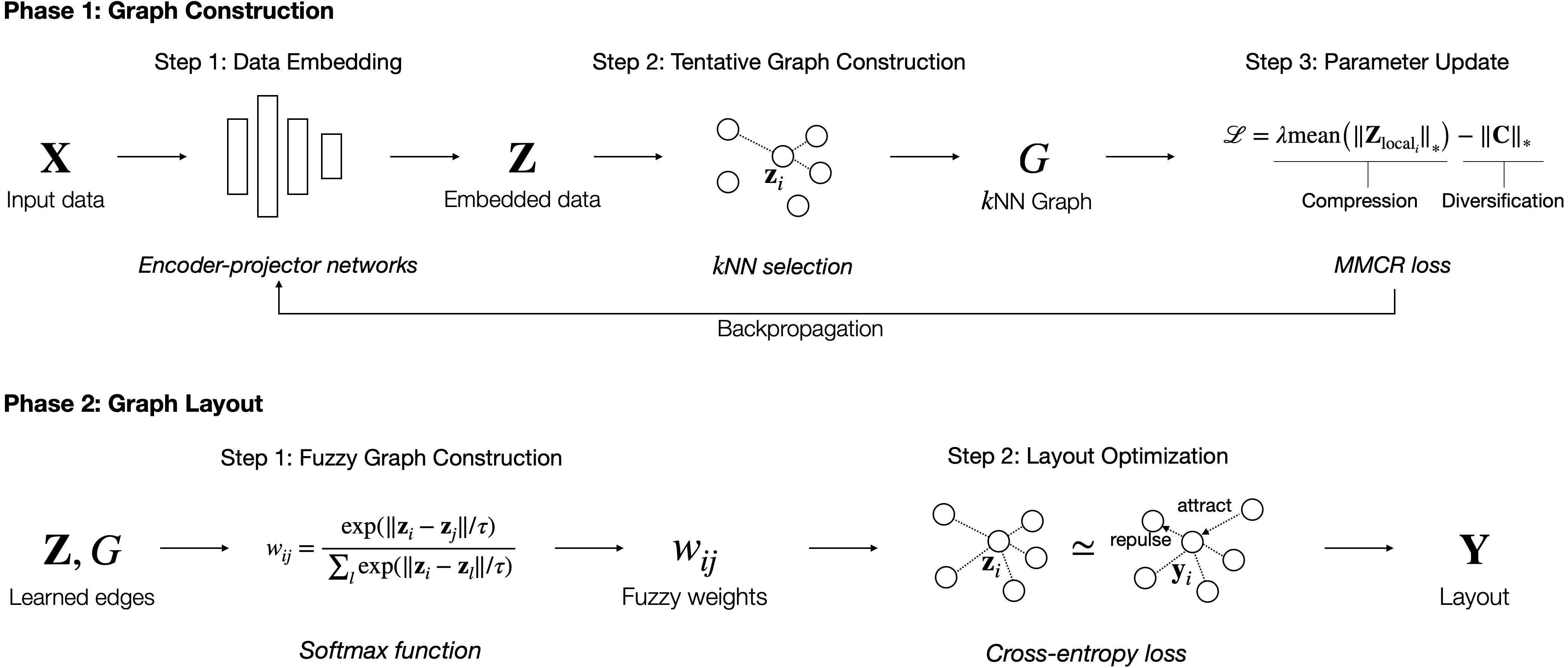}
    \caption{The MAPLE pipeline, consisting of two phases: (1) graph construction and (2) graph layout. 
    Phase 1 learns a graph representation of the input data through three steps that integrate SSL with MMCRs. 
    Phase 2 performs the cross-entropy-based layout optimization, following UMAP.
    MAPLE's core contribution lies in Phase 1.}
    \label{fig:pipeline}
\end{figure*}

\subsection{Motivation} \label{sec:maple_motivation}
High-dimensional data introduce fundamental challenges for DR methods with graph-based modeling, such as $t$-SNE~\cite{maaten2008visualizing}, UMAP~\cite{mcinnes_umap_2020}, and PaCMAP~\cite{wang2021understanding}.
First, as dimensionality increases, conventional distance metrics (e.g., Euclidean and cosine distances) become more prone to the curse of dimensionality---pairwise distances measured in the high-dimensional space become more sensitive to noise within a data distribution~\cite{beyer1999nearest}. 
This makes it difficult to distinguish true neighbors from unrelated points when generating neighbor graphs from high-dimensional data.
Second, many high-dimensional datasets (e.g., images or single-cell data in genomics) vary along curved low-dimensional manifolds defined by nonlinear generative factors such as pose or cell state.
In these cases, close points in the high-dimensional space may be far apart along the underlying manifold and, therefore, may produce neighbor graphs where semantically distinct regions are connected.
A recent DR method, LocalMAP~\cite{wang2025dimension}, is designed to mitigate this issue in PaCMAP~\cite{wang2021understanding}. 
LocalMAP dynamically reweights neighbor edges and resamples neighbors during optimization.
However, this approach relies on heuristic adjustments and does not resolve the broader issue of noisy variance and manifold-level distortions when constructing neighborhood graphs.

UMAP models the underlying manifold as a fuzzy graph derived from a $k$NN graph that is constructed based on distances in high-dimensional space.
Due to the above two-fold characteristics of high-dimensional data, the fuzzy graph can cause semantic mismatches, where points with no meaningful relationships are connected only because of distances in high-dimensional space rather than the underlying manifold geometry.
As discussed in \cref{sec:umap}, these semantic mismatches propagate into the final layout.
This issue often manifests as loss of subcluster details in UMAP's outcomes when intra-cluster variation is not adequately represented, and as noisy or irregular cluster boundaries when neighborhoods contain points belonging to different clusters.
Therefore, with MAPLE, we aim to refine the graph construction phase of UMAP (see Phase 1 in \cref{sec:umap}).

\subsection{Overview of MAPLE} 
MAPLE enhances graph-based modeling of the underlying manifold through SSL.
\cref{fig:intro} illustrates a conceptual process by which MAPLE produces a concise graph representation of the manifold.
Building on UMAP's pipeline, MAPLE shifts graph construction from a fixed-distance space (e.g., defined by Euclidean distances) into a learned embedding space.
This shift is enabled by incorporating MMCRs during graph construction. 
MMCRs introduce two complementary forces: (1) a compression force that reduces variance within local neighborhoods, flattening noisy or curved regions to make substantial structures (e.g., subclusters) more apparent; and (2) a diversification force that increases variance across dissimilar regions, reducing edges that connect dissimilar points and clarifying boundaries. 
The learned graph is then laid out in a low-dimensional space using UMAP's cross-entropy optimization.

\subsection{Phase 1: Self-supervised Graph Construction}

MAPLE's graph construction is performed through three steps, as illustrated in Phase 1 of \cref{fig:pipeline}. 
This phase represents the core contribution of MAPLE, where we integrate the MVSSL scheme into the DR process and incorporate MMCRs into the manifold modeling objective.

\subsubsection{Step 1: Data Embedding}

In the first step, we embed the original high-dimensional data into a lower-dimensional space (e.g., 128 dimensions) using simple neural networks. This reduces the influence of noisy data points, thereby improving both training stability and the quality of the learned graph.

Following prior work in self-supervised learning~\cite{grill2020bootstrap,kalantidis2021tldr,zbontar_barlow_2021}, MAPLE employs \textit{encoder and projector networks}. 
The encoder network, $\EncoderNetwork$, consists of two fully connected hidden layers (default: 512 and 2,048 nodes), while the projector network, $\ProjectorNetwork$, consists of two layers (512 and 128 nodes).
Together, these networks transform the original high-dimensional data, $\OriginalData$, into embedded data, $\EmbedData \in \mathbb{R}^{\nPoints \times \nEmbedDimensions}$ where $\nEmbedDimensions = 128$ under the default setting (i.e., $\EmbedData = \ProjectorNetwork ( \EncoderNetwork ( \OriginalData ) )$).
During this transformation, each embedded data point is normalized to a unit vector to eliminate scaling differences when computing similarities.
This unit-vector constraint is also consistent with the original implementation of MMCRs~\cite{yerxa2023mmcr}.

We apply this default architecture across datasets as a generic, lightweight feature extractor.
For high-dimensional inputs ($D > 512$), the first layer provides a moderate compression that helps filter out noise and stabilizes subsequent training.
For lower-dimensional inputs ($D < 512$), although the first layer could technically be reduced in size, we retain this expanded capacity to maintain a consistent and standardized architecture across datasets.
\textcolor{minor}{Alternatively, the architecture could be scaled (e.g., linearly) with the input dimensionality; exploring such scaling strategies is a worthwhile direction for future work.}
Although more complex architectures such as ResNet~\cite{he2016deep} could be employed, we deliberately adopt this minimal encoder--projector design as an efficient and broadly applicable approach for high-dimensional data.

\subsubsection{Step 2: Tentative Graph Construction}
\label{sec3:tentative_graph}

The second step constructs a tentative graph, $\Graph$, from the current embedded data, $\EmbedData$.
We first build a $k$NN graph as $\Graph$ based on $L^2$ distances in the embedding space.
Unlike UMAP, MAPLE iteratively refines $\Graph$ by updating the neural network parameters of $\EncoderNetwork$ and $\ProjectorNetwork$ with MMCRs, as described in Step 3.

The neighbors in the tentative graph can be viewed as noisy ``self-references''.
We treat neighbors as similar data points and non-neighbors as dissimilar ones to enable self-supervised learning in Step~3. 
In the context of MVSSL, these neighbor data points can be regarded as \textit{augmented views} of each data point, used to learn invariances.
In our case, a data point and its neighbors act as multiple views reflecting the same local structure.
However, unlike typical augmentation scenarios (e.g., creating augmented images via rotation), these views may not align perfectly with the same structure.
Since the networks are initially untrained, early-epoch neighborhoods can still reflect noisy structure inherited from $\OriginalData$.
We therefore refine these neighborhoods through regularization with MMCRs.

\subsubsection{Step 3: MMCR Loss Computation and Parameter Update}

In MAPLE, we reinterpret the objective of SSL through MMCRs~\cite{yerxa2023mmcr}, specifically in the context of neighbor graph learning for DR.
MMCRs aim to produce \textit{efficient} data embeddings for downstream classification tasks by maximizing the linear separability of local manifolds, which often correspond to different classes, in the embedding space.
Within neighbor graph learning, MMCRs refine the embedding space to better preserve neighborhood relationships along the underlying manifold.

MMCRs consider two types of variance to improve the linear separability of local manifolds: (1) \textit{between-variance}, the variance of samples across different local manifolds, and (2) \textit{within-variance}, the variance of samples within the same local manifold. 
Linear separability improves when between-variance increases and within-variance decreases.

\textbf{Between-variance.} \quad 
As discussed in \cref{sec3:tentative_graph}, we treat each embedded data point and its neighborhoods as views of the same structure.
In other words, we assume these points are samples from the same local manifold.
Let $\Members_i$ be the set of indices of the $i$th data point and its $k$ nearest neighbors.
For the local manifold around a data point $\MakeLowercase{\EmbedData}_i$, the centroid of the manifold, $\MakeLowercase{\Centroid}_i$, can be approximated as: $\MakeLowercase{\Centroid}_i = \sum_{j \in \Members_i}  \MakeLowercase{\EmbedData}_j / (k + 1)$.
The between-variance is then defined as: 
\begin{equation} \label{eq:global_mmcr}
    \mathrm{Var}_\mathrm{between} := \norm{\Centroid}_*
\end{equation}
where $\Centroid = [\MakeLowercase{\Centroid}_1 \cdots \MakeLowercase{\Centroid}_\nPoints]^\top \in \mathbb{R}^{\nPoints \times \nEmbedDimensions}$ and $\norm{\cdot}_*$ denotes the nuclear norm operator, with $\norm{\Centroid}_* = \mathrm{trace}(\sqrt{\Centroid \Centroid^\top})$.
Intuitively, the nuclear norm measures the total variance of $\Centroid$ across all principal directions. 
$\norm{\Centroid}_*$ is large when the centroids spread across many orthogonal directions, and small when the centroids nearly lie in a low-dimensional subspace (e.g., a plane).

Maximizing the between-variance encourages the centroids of local manifolds to spread across multiple orthogonal directions. 
In other words, it promotes variance to be distributed more evenly across multiple principal components, rather than collapsing into a few dominant axes. 
Geometrically, this pushes centroids to form large pairwise angles on the unit sphere, suppressing concentration in a single direction that would otherwise create hub nodes in the neighborhood graph.

\textbf{Within-variance.} \quad 
Let $\EmbedData_{\mathrm{local}_i}$ be the matrix consisting of the $i$-th data point in $\EmbedData$ and its $k$ nearest neighbors (i.e., $\EmbedData_{\mathrm{local}_i} \in \mathbb{R}^{(k+1) \times \nEmbedDimensions}$).
Then, similar to the between-variance, the within-variance of the local manifold around each point $\MakeLowercase{\EmbedData}_i$ is defined as:
\begin{equation}  \label{eq:local_mmcr}
    \mathrm{Var}_\mathrm{within} := \norm{\EmbedData_{\mathrm{local}_i}}_*
\end{equation}

Minimizing the within-variance encourages samples from each local manifold to align in a lower-dimensional subspace, ideally approximating a flat tangent plane. 
The nuclear norm is upper-bounded by the Frobenius norm~\cite{peng2016connections} scaled by the square root of the rank, i.e., $\norm{\EmbedData_{\mathrm{local}_i}}_* \leq \sqrt{rank(\EmbedData_{\mathrm{local}_i})}\norm{\EmbedData_{\mathrm{local}_i}}_F$. 
Thus, minimizing the within-variance suppresses variance orthogonal to the leading singular directions, making each neighborhood more compact and low-rank.
As a result, semantic neighbors are pulled closer together, while irrelevant neighbors are naturally down-weighted.

\textbf{MMCR loss function.} \quad
Using the between- and within-variances of local manifolds, we define the MMCR loss as:
\begin{equation}
    \label{eq:mmcr_loss}
    \Loss = \lambda \underset{i \in \{1, \cdots, \nPoints \}}{\mathrm{mean}} \left( \norm{\EmbedData_{\mathrm{local}_i}}_* \right) - \norm{\Centroid}_* 
\end{equation}
where $\lambda > 0$ is a trade-off hyperparameter.
This loss balances two complementary forces: the first term acts as a compression force, shrinking the size of local manifolds, while the second term acts as a diversification force, encouraging separation among different local manifolds.
$\lambda$ controls the relative strength of these forces.
MAPLE optimizes this loss by updating the parameters $\theta$ of the encoder and projector networks, ensuring that the resulting embedding $\EmbedData$ minimizes $\Loss$ (i.e., solving $\arg \min_\theta \Loss$).

\textbf{Efficient optimization.} \quad
Although Steps 1--3 could, in theory, be repeated at every update, we adopt a computationally efficient optimization strategy.
First, we use mini-batch gradient descent, computing the loss over batches of data.
Second, we reuse the constructed graph across updates in Steps 1 and 3, and only reconstruct it every fixed number of epochs (20 by default).

\textbf{Alternative designs.} \quad
MAPLE uses the encoder--projector networks and MMCRs. 
We also explored several alternative designs during development.
For example, we tested multiple different self-supervised learning approaches such as Barlow Twins~\cite{zbontar_barlow_2021}.
Our supplementary materials provide detailed discussions on tested designs.

\subsection{Phase 2: Graph Layout}
From the outcomes of Phase 1, MAPLE constructs a fuzzy graph and then performs layout optimization, as illustrated in \cref{fig:pipeline}.

\subsubsection{Step 1. Fuzzy Graph Construction}

Given the learned graph, $\Graph$, and the embedded data, $\EmbedData$, MAPLE computes fuzzy weights for each edge using a softmax function:
\begin{equation}
    \EdgeWeight = \frac{\exp({-\norm{\MakeLowercase{\EmbedData}_i - \MakeLowercase{\EmbedData}_j} / \tau}) }{\sum \limits_{l \in i\mathrm{'s\ neighbors}} \exp({-\norm{\MakeLowercase{\EmbedData}_i - \MakeLowercase{\EmbedData}_l} / \tau})}
\end{equation}
where $\tau$ is a softmax temperature parameter and $\sum_j\EdgeWeight = 1$. 
$\EdgeWeight$ defines a probability distribution over $i$'s neighbors: closer neighbors (higher similarity) receive higher weights, while the softmax normalizes across the neighborhood, balancing the attractive forces from different neighbors.
MAPLE's fuzzy weight computation differs from UMAP's original one (\cref{eq:umap_weight}), which is designed to eliminate weak neighbor connections.
Since the learned graph in MAPLE has already removed such connections, applying UMAP's procedure would result in unnecessary elimination.

\subsubsection{Step 2: Layout Optimization}
For layout optimization, we adopt UMAP's original approach.
The layout, $\Layout$, is initialized using a spectral layout and then optimized based on the cross-entropy loss between pairwise similarities in the embedded space (128D by default) and the layout space (typically 2D or 3D).
We preserve all aspects of UMAP's optimization procedure, including negative sampling, learning rate scheduling, and stochastic updates. 
This setting ensures that MAPLE's improvements primarily arise from the self-supervised graph construction rather than modifications to the layout optimization.

In many cases (e.g., \cref{fig:intro}, \cref{fig:case}(a)), MAPLE produces locally linear, skeleton-like motifs in the final layout.
This behavior can be interpreted as a geometric consequence of optimizing manifold capacity in \cref{eq:mmcr_loss}.
Since the nuclear norm jointly reflects manifold radius and intrinsic dimensionality~\cite{yerxa2023mmcr}, minimizing this term suppresses variance along minor directions within local neighborhoods. 
As a result, local structures are encouraged to align with their principal axes, approximating low-rank subspaces rather than high-dimensional ellipsoids.
Compared to UMAP, which often yields a series of blob-like clusters, this process can produce layouts composed of concatenated, locally flattened segments.

\vspace{5pt}
\noindent\textbf{Implementation} \quad
We implemented MAPLE in Python 3, using NumPy/SciPy~\cite{virtanen2020scipy} for matrix calculations and scikit-learn~\cite{pedregosa2011scikit} for $k$NN graph construction.
The neural networks are built with PyTorch~\cite{paszke2019pytorch}.
For the MMCR loss, we adopted the implementation by the original authors~\cite{yerxa2023mmcr}.
For layout optimization, we employed the official UMAP implementation~\cite{mcinnes_umap_2020}.
Additional implementation details, including neural network settings, are provided in the supplementary materials.

\vspace{5pt}
\noindent\textbf{Summary of main hyperparameters} \quad
MAPLE introduces a small set of hyperparameters: (1) the embedding dimension $\nEmbedDimensions$, (2) the number of neighbors $k$ for graph construction, (3) the trade-off parameter $\lambda$ in the MMCR loss, and (4) the softmax temperature $\tau$ for the fuzzy weight computation.
Based on our experiments across multiple datasets, we observed that MAPLE is relatively robust to moderate variations in $\nEmbedDimensions$ and $\tau$, due to the stability of the UMAP cross-entropy objective. 
Therefore, we consider $k$ and $\lambda$ to be the primary hyperparameters governing MAPLE's behavior.
Both UMAP and MAPLE require the neighborhood size hyperparameter $k$.
The remaining main parameter $\lambda$ in MAPLE plays an analogous role to \texttt{min\_dist} in UMAP, controlling the balance between diversification and compression forces.
However, a key distinction is that while \texttt{min\_dist} imposes a static constraint on inter-point distances, the MAPLE objective is grounded in manifold capacity theory and adjusts the balance between compression and diversification during optimization.
By optimizing the nuclear norm, the objective adaptively influences the 
spacing and density of local manifolds in the embedding space.
Our parameter sensitivity analysis (detailed in \cref{sec:parameter}) further shows that MAPLE remains robust across a wide range of $\lambda$ and varying $k$ values.

\section{Results}
\label{sec:results}
\begin{figure*}[ht!]
\centering
    \includegraphics[width=1\linewidth]{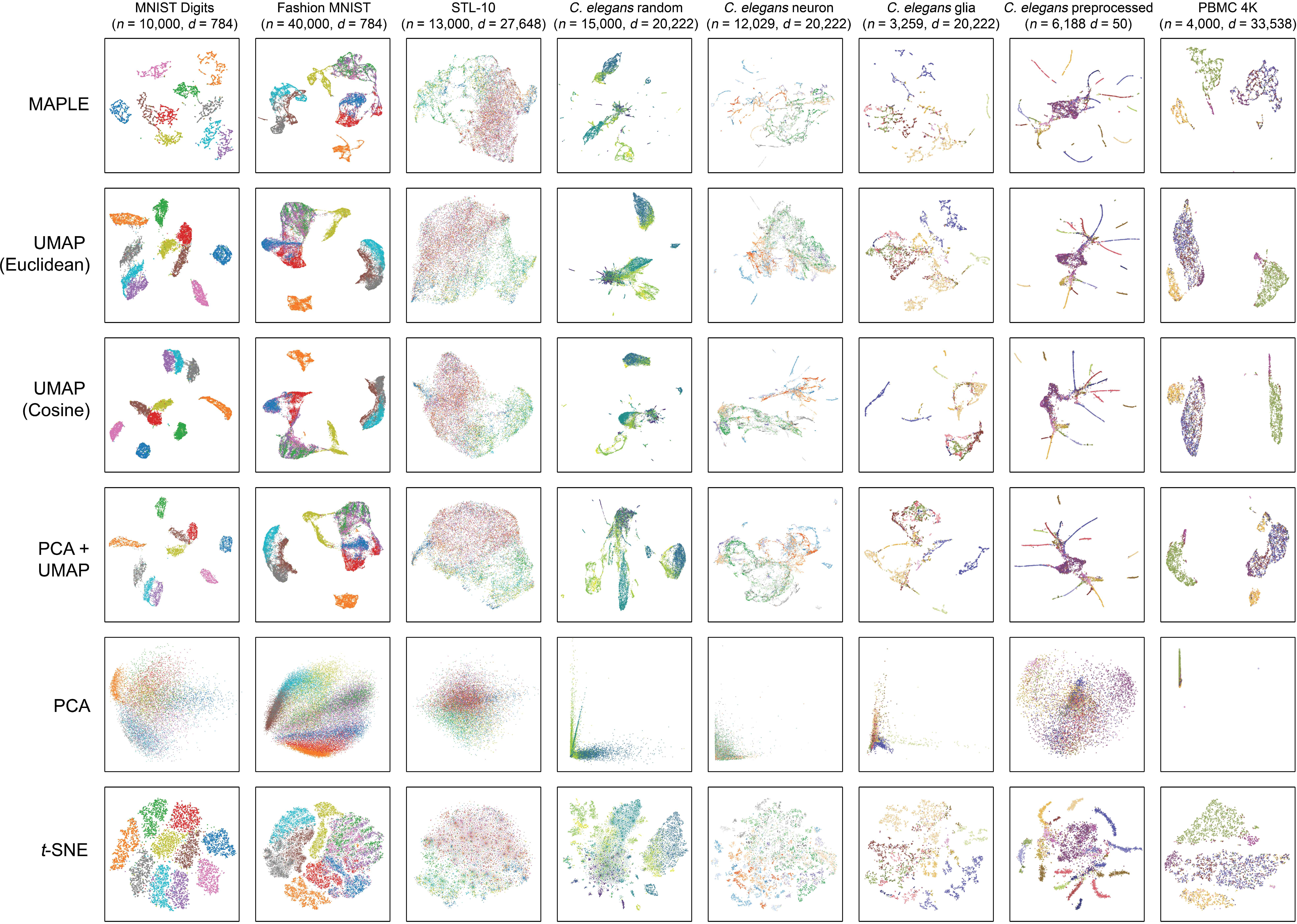}
    \caption{Layouts of image (MNIST~\cite{deng2012mnist}, Fashion-MNIST~\cite{xiao2017fashion}, STL-10~\cite{coates2011analysis}) and single-cell datasets (\textit{C. elegans} subsets~\cite{packer2019lineage} and PBMC 4K) using MAPLE and baseline methods. MAPLE preserves both global structure and local neighborhoods more consistently across datasets.}
    \label{fig:comp}
\end{figure*}

\begin{table*}[t]
\centering
\caption{MAPLE vs UMAP across local, global, clustering, boundary, and perceptual measures (higher is better $\uparrow$, lower is better $\downarrow$). 
} 
\label{tab:metrics}
\scriptsize
\setlength{\tabcolsep}{3pt}
\resizebox{\textwidth}{!}{%
\begin{tabular}{l
rr rr rr rr rr rr rr rr}
\toprule
& \multicolumn{2}{c}{\textbf{MNIST Digits}}
& \multicolumn{2}{c}{\textbf{Fashion MNIST}}
& \multicolumn{2}{c}{\textbf{STL-10}}
& \multicolumn{2}{c}{\textbf{\textit{C. el} Random}}
& \multicolumn{2}{c}{\textbf{\textit{C. el} Neuron}}
& \multicolumn{2}{c}{\textbf{\textit{C. el} Glia}}
& \multicolumn{2}{c}{\textbf{\textit{C. el} Preprocessed}}
& \multicolumn{2}{c}{\textbf{PBMC 4K}}\\
\cmidrule(r){2-3}\cmidrule(r){4-5}\cmidrule(r){6-7}\cmidrule(r){8-9}\cmidrule(r){10-11}\cmidrule(r){12-13}\cmidrule(r){14-15}\cmidrule(r){16-17}
\textbf{Measure} & MAPLE & UMAP & MAPLE & UMAP & MAPLE & UMAP & MAPLE & UMAP & MAPLE & UMAP & MAPLE & UMAP & MAPLE & UMAP & MAPLE & UMAP \\
\midrule
\multicolumn{17}{l}{\textit{Local}}\\
Neighborhood hit (core) $\uparrow$ & \textbf{0.922} & 0.911 & \textbf{0.761} & 0.715 & \textbf{0.244} & 0.212 & \textbf{0.554} & 0.450 & \textbf{0.626} & 0.561 & \textbf{0.750} & 0.718 & \textbf{0.756} & 0.751 & \textbf{0.598} & 0.576 \\
$k$NN classification accuracy (core) $\uparrow$ & \textbf{0.813} & 0.794 & \textbf{0.818} & 0.778 & \textbf{0.302} & 0.260 & \textbf{0.650} & 0.554 & \textbf{0.660} & 0.555 & \textbf{0.825} & 0.790 & \textbf{0.859} & 0.855 & \textbf{0.678} & 0.601 \\
Distance consistency $\uparrow$ & \textbf{0.913} & 0.911 & 0.660 & \textbf{0.670} & 0.203 & \textbf{0.224} & \textbf{0.341} & 0.247 & \textbf{0.307} & 0.227 & \textbf{0.424} & 0.418 & 0.519 & \textbf{0.606} & \textbf{0.255} & 0.245 \\
\addlinespace[2pt]
\multicolumn{17}{l}{\textit{Global}}\\
CH $\uparrow$ & 9282.101 & \textbf{11480.073} & 25828.332 & \textbf{36460.773} & \textbf{643.450} & 522.385 & \textbf{521.646} & 471.087 & \textbf{557.774} & 517.456 & \textbf{463.124} & 457.781 & 678.806 & \textbf{1205.347} & 560.435 & \textbf{1262.673} \\
DB $\downarrow$ & \textbf{0.926} & 0.971 & 2.055 & \textbf{1.947} & \textbf{16.232} & 34.250 & \textbf{15.664} & 19.263 & \textbf{8.131} & 9.393 & 3.719 & \textbf{3.170} & 5.360 & \textbf{4.245} & \textbf{5.119} & 6.050 \\
\addlinespace[2pt]
\multicolumn{17}{l}{\textit{Clustering (k-means)}}\\
ARI $\uparrow$ & \textbf{0.810} & 0.752 & \textbf{0.493} & 0.431 & \textbf{0.072} & 0.063 & \textbf{0.237} & 0.183 & \textbf{0.174} & 0.159 & \textbf{0.365} & 0.351 & 0.343 & \textbf{0.436} & \textbf{0.337} & 0.294 \\
AMI $\uparrow$ & 0.792 & \textbf{0.813} & \textbf{0.651} & 0.623 & \textbf{0.158} & 0.138 & \textbf{0.484} & 0.469 & \textbf{0.434} & 0.397 & \textbf{0.540} & 0.531 & 0.547 & \textbf{0.571} & \textbf{0.524} & 0.502 \\
NMI $\uparrow$ & \textbf{0.793} & 0.779 & \textbf{0.655} & 0.607 & \textbf{0.164} & 0.141 & \textbf{0.483} & 0.477 & \textbf{0.424} & 0.412 & \textbf{0.545} & 0.532 & 0.567 & \textbf{0.614} & \textbf{0.530} & 0.507 \\
V-measure $\uparrow$ & \textbf{0.832} & 0.779 & \textbf{0.649} & 0.607 & \textbf{0.159} & 0.139 & \textbf{0.515} & 0.476 & \textbf{0.428} & 0.408 & \textbf{0.548} & 0.522 & 0.595 & \textbf{0.615} & \textbf{0.543} & 0.510 \\
\addlinespace[2pt]
\multicolumn{17}{l}{\textit{Clustering (DBSCAN)}}\\
ARI $\uparrow$ & \textbf{0.711} & 0.405 & \textbf{0.334} & 0.219 & - & - & \textbf{0.203} & 0.061 & \textbf{0.157} & 0.074 & \textbf{0.345} & 0.078 & \textbf{0.166} & 0.056 & \textbf{0.447} & 0.400 \\
AMI $\uparrow$ & \textbf{0.794} & 0.689 & \textbf{0.596} & 0.532 & - & - & \textbf{0.514} & 0.296 & \textbf{0.499} & 0.348 & \textbf{0.552} & 0.379 & \textbf{0.525} & 0.322 & \textbf{0.584} & 0.568 \\
NMI $\uparrow$ & \textbf{0.795} & 0.690 & \textbf{0.596} & 0.532 & - & - & \textbf{0.528} & 0.302 & \textbf{0.511} & 0.355 & \textbf{0.563} & 0.384 & \textbf{0.534} & 0.328 & \textbf{0.588} & 0.571 \\
V-measure $\uparrow$ & \textbf{0.795} & 0.690 & \textbf{0.596} & 0.532 & - & - & \textbf{0.528} & 0.302 & \textbf{0.511} & 0.355 & \textbf{0.563} & 0.384 & \textbf{0.534} & 0.328 & \textbf{0.588} & 0.571 \\
\addlinespace[2pt]
\multicolumn{17}{l}{\textit{Boundary}}\\
Label trustworthiness $\uparrow$ & \textbf{0.846} & 0.823 & \textbf{0.523} & 0.431 & \textbf{-0.511} & -0.576 & \textbf{0.109} & -0.099 & \textbf{0.255} & 0.124 & \textbf{0.504} & 0.438 & \textbf{0.513} & 0.503 & \textbf{0.198} & 0.152 \\
Hausdorff distance $\uparrow$ & \textbf{5.112} & 2.938 & \textbf{5.211} & 3.077 & \textbf{0.792} & 0.574 & \textbf{8.898} & 2.512 & \textbf{6.583} & 4.074 & \textbf{7.761} & 4.800 & \textbf{7.994} & 5.674 & \textbf{11.049} & 7.430 \\
\addlinespace[2pt]
\multicolumn{17}{l}{\textit{Perceptual}}\\
2AFC accuracy $\uparrow$ & \textbf{0.953} & 0.949 & 0.898 & \textbf{0.904} & \textbf{0.612} & 0.602 & \textbf{0.861} & 0.848 & \textbf{0.828} & 0.809 & 0.840 & \textbf{0.844} & 0.853 & \textbf{0.893} & 0.869 & \textbf{0.870} \\
Pairwise AUC (ROC--AUC) $\uparrow$ & 0.953 & \textbf{0.954} & 0.908 & \textbf{0.918} & \textbf{0.610} & 0.603 & \textbf{0.795} & 0.747 & \textbf{0.796} & 0.792 & \textbf{0.845} & 0.827 & 0.868 & \textbf{0.895} & \textbf{0.881} & 0.879 \\
\bottomrule
\vspace{-4pt}
\end{tabular}}
\raggedright
Note: Best scores per dataset are in \textbf{bold}. The \textit{C. elegans} (\textit{C. el}) datasets (Random, Neuron, Glia) are evaluated against fine-grained biological subtype annotations rather than broad cell classes. Clustering scores for STL-10 are not included because the density-based clustering algorithm DBSCAN identified only a single cluster for both MAPLE and UMAP.
\end{table*}

We evaluate MAPLE on eight datasets with multiple classes or subtypes, representative of real-world high-dimensional analysis challenges. 
\cref{fig:comp} shows layouts generated by MAPLE and five other DR methods. 
\cref{tab:metrics} presents quantitative results across 17 evaluation measures on the same datasets (in the same order as~\cref{fig:comp}), comparing MAPLE with UMAP using a Euclidean input-space metric.
The corresponding layouts can be found in the first two rows of~\cref{fig:comp}.

\textbf{Datasets.} \quad 
We use datasets from two representative domains that involve very high-dimensional data and commonly utilize DR for their analyses: image recognition and transcriptomics.
\begin{description}
[font=\normalfont,topsep=2pt,itemsep=1pt,leftmargin=2pt]
    \item[Image datasets:]\leavevmode
    \begin{enumerate}[topsep=0pt]
        \item MNIST handwritten digits test set~\cite{deng2012mnist}: $\nPoints {=} 10{,}000$, $\nDimensions {=} 784$
        \item A subset of Fashion-MNIST~\cite{xiao2017fashion}: $\nPoints {=} 40{,}000$, $\nDimensions {=} 784$
        \item Labeled images from STL-10~\cite{coates2011analysis}: $\nPoints {=} 13{,}000$, $\nDimensions {=} 27{,}648$
    \end{enumerate}
    \vspace{2pt}
    \item[Single-cell RNA sequencing (scRNA-seq) datasets:]\leavevmode
    \begin{enumerate}[topsep=0pt]
        \item A random subset of \textit{C. elegans}~\cite{packer2019lineage}: $\nPoints {=} 15{,}000$, $\nDimensions=20{,}222$
        \item Neuron subset from \textit{C. elegans}: $\nPoints {=} 12{,}029$, $\nDimensions=20{,}222$
        \item Glia subset from \textit{C. elegans}: $\nPoints {=} 3{,}259$, $\nDimensions=20{,}222$
        \item Preprocessed subset of \textit{C. elegans}: $\nPoints {=} 6{,}188$, $\nDimensions=50$
        \item PBMC 4K from 10x Genomics~\cite{10xgenomics}, annotated with the Cellarium Cell Annotation Service~\cite{cell_annotation_service}: $\nPoints {=} 4{,}000$, $\nDimensions=33{,}538$ 
    \end{enumerate}
\end{description}
The preprocessed subset of \textit{C. elegans} follows the pipeline recommended in Monocle’s documentation~\cite{monocle3trajectories}. 
We include this dataset to assess MAPLE's performance when strong preprocessing has already been applied.

\subsection{Quantitative Evaluation}
Quantitative evaluation of DR layouts is inherently challenging, as there is no universally accepted ``ground truth'' for high-dimensional structure, and different measures emphasize different aspects of layout quality~\cite{espadoto2019toward, jung2025ghostumap2, machado2025necessary, jeon2023classes}.
To perform quantitative evaluation, it is therefore necessary to select or design quality measures for DR layouts~\cite{jeon2023zadu,jeon2025unveiling}.
While existing measures provide valuable insights, selecting appropriate ones remains challenging due to inherent trade-offs in the structural aspects they capture.
A large class of existing measures heuristically quantifies how well neighborhood relationships in high-dimensional space are preserved in the resulting layout. 
For instance, the Kullback--Leibler divergence~\cite{hinton2002stochastic} compares neighborhood probability distributions, while trustworthiness and continuity~\cite{venna2006local} penalize neighbors that are gained or lost compared to the input $k$NNs.
ZADU~\cite{jeon2023zadu}, a library compiling various quality measures, includes 20 measures, 15 of which fall into this class. 

Although widely adopted as standard evaluation criteria for new DR methods~\cite{atzberger2024large, jung2025ghostumap2}, these measures capture only limited aspects of layout quality and face two critical issues.
First, many rely on $k$NN graphs constructed with distance metrics; however, as discussed in \cref{sec:umap} and~\cref{sec:maple_motivation}, conventional $k$NN graphs do not necessarily reflect intrinsic manifold structure~\cite{beyer1999nearest, bergman2020deep, bohm2022unsupervised, wang2025dimension}.
Second, these measures emphasize statistical interpretability rather than geometric or informational expressiveness. 
They quantify whether certain local statistics (e.g., distances among neighbors) are preserved.
However, as highlighted in SSL research~\cite{bear2020learning, galvez2023role}, raw data often contain noise, and such statistics may not reflect whether embeddings and layouts provide faithful geometric representations of the manifold (e.g., curvature, geodesics~\cite{moon_PHATE_2019}).
MAPLE is designed to address data with noisy local manifolds, and the layout quality should be judged by the ability to reveal semantically meaningful structures.

Given these limitations, we use class labels as a proxy for ground truth semantic structure in high-dimensional space.
Accordingly, we use label-based measures to quantify layout quality.
Jeon et al.~\cite{jeon2023classes} argue that label-based measures can be unreliable as classes do not always correspond to clusters in the original space.
This concern is mitigated in our evaluation by our choice of datasets, where classes are well-defined.

As a result, we assembled a set of 17 label-based quality measures suitable for comprehensive evaluation.
Specifically, this set consists of five groups:  
\begin{itemize}
    \item \textit{Local measures}, including neighborhood hit, $k$NN classification accuracy~\cite{mcinnes_umap_2020}, and distance consistency, which measure label agreement at the neighborhood and centroid level~\cite{paulovich2008least, jeon2023zadu};  
    \item \textit{Global measures}, including the Calinski--Harabasz (CH)~\cite{calinski1974dendrite} and Davies--Bouldin (DB)~\cite{vergani2018soft} scores, both of which assess the quality of class margins (with higher CH and lower DB indicating clearer separation)~\cite{jeon2023zadu};  
    \item  \textit{Clustering measures}, obtained by applying $k$-means~\cite{hamerly2003learning} and DBSCAN~\cite{ester1996density} on the 2D layouts and reporting four scores---adjusted Rand index (ARI), adjusted mutual info score (AMI), normalized mutual info score (NMI), and V-measure---to quantify alignment with ground-truth classes~\cite{vinh2009information, jeon2023zadu};  
    \item \textit{Boundary measures}, which evaluate boundary quality at both local and global levels; and
    \item \textit{Perceptual measures}, adapted from perceptual similarity measures in computer vision~\cite{zhang2018unreasonable} to mimic human judgments.  
\end{itemize}

\noindent For consistency, the local, global, and clustering measures in \cref{tab:metrics} were reimplemented from ZADU~\cite{jeon2023zadu} in PyTorch. 
For all quantitative comparisons, UMAP was evaluated using neighborhood sizes matched to MAPLE, while all other UMAP parameters followed default settings. Additional implementation details and parameter settings are provided in the supplementary materials.

For the \textit{local boundary} measure, we modify the classical trustworthiness score~\cite{venna2006local} into a local label trustworthiness measure. For each point, neighbors with mismatched labels are penalized more strongly when they appear among the closest neighbors. 
The final score is the average across all points, reflecting the cleanliness of class boundaries in the layout.
To measure \textit{boundary globally}, we compute a quantile Hausdorff distance~\cite{huttenlocher2002comparing, taha2015metrics} between clusters. 
For each pair of classes, we measure how close points from one class are to the other, with the focus on the nearest cross-class neighbors. Instead of taking only the single closest pair, which can be noisy, we use the percentile of these nearest-neighbor distances and then obtain the median of these values across all class pairs.

We design \textit{perceptual} measures inspired by perceptual similarity tests in computer vision~\cite{zhang2018unreasonable}, which mimic how people judge whether two images look alike.
We adopt the same sampling-based idea for evaluating the layouts perceptually.
We randomly sample reference points from the layout produced by MAPLE or UMAP, and compare distances to same-class and different-class neighbors. 
Based on Zhang et al.~\cite{zhang2018unreasonable}, we define a two-alternative forced choice (2AFC) accuracy measure, which counts how often a reference is closer to a same-class point than to a different-class point.
Similar to the Just Noticeable Difference (JND) test~\cite{zhang2018unreasonable}, we also compute the area under the receiver operating characteristic curve (ROC--AUC) score~\cite{bradley1997use}, which corresponds to sampling many pairs of points from the layout, recording their pairwise distances, and labeling them as same-class or different-class. 
The ROC curve captures how well pairwise distance discriminates between same-class and different-class pairs across thresholds, and the resulting AUC provides a single reliability score. 
Unlike the original perceptual measure that relies on deep neural network features (e.g., from AlexNet~\cite{krizhevsky2012imagenet}) to approximate human visual perception, we sample points on the DR results and check them with ground-truth labels.

\begin{table}[t]
\centering
\caption{Per-class \textit{k}NN classification accuracy scores (with $k = 10$).
}
\vspace{-5pt}
\label{tab:combined_results}
\resizebox{\columnwidth}{!}{%
\setlength{\tabcolsep}{2pt}
\begin{tabular}{lrrrrrrrrrr} 
\toprule
\multicolumn{11}{c}{\textbf{Fashion MNIST}} \\
\midrule
 & T-shirt & Trouser & Pullover & Dress & Coat & Sandal & Shirt & Sneaker & Bag & Ankle Boot \\
\midrule
MAPLE
& \textbf{0.838} & \textbf{0.960} & \textbf{0.697} & \textbf{0.869} & \textbf{0.676}
& \textbf{0.913} & \textbf{0.442} & \textbf{0.907} & \textbf{0.960} & \textbf{0.923} \\
UMAP
& 0.830 & 0.951 & 0.643 & 0.839 & 0.576
& 0.865 & 0.372 & 0.854 & 0.939 & 0.921 \\
\midrule
\midrule
\multicolumn{11}{c}{\textbf{\textit{C. elegans} Neuron}} \\
\midrule
 & ADF & ADF\_AWB & AFD & ASE & ASE\_p & ASK & ASK\_p & OLQ & OLQ\_gp & OLQ\_p \\
\midrule
MAPLE
& \textbf{0.718} & \textbf{0.588} & \textbf{0.825} & \textbf{0.745} & \textbf{0.772} & \textbf{0.764} & \textbf{0.567} & 0.717 & 0.514 & \textbf{0.671} \\
UMAP
& 0.588 & 0.343 & 0.696 & 0.585 & 0.651 & 0.506 & 0.440 & \textbf{0.720} & \textbf{0.590} & 0.578 \\

\bottomrule
\end{tabular}%
}
\par\vspace{4pt}
\parbox{0.98\linewidth}{
  \raggedright\footnotesize
  \setlength{\parskip}{0pt}
  \setlength{\baselineskip}{8.5pt}
  Note: Additional results with varying $k$ values and full per-class scores are provided in the supplementary materials.
}
\vspace{-2pt} 
\end{table}

\textbf{Observations on core quantitative measures.}\quad
We focus here on \textit{neighborhood hit} and \textit{$k$NN classification accuracy}, as these measures are directly related to assessing the quality of the learned neighborhood structure in MAPLE.
As shown in \cref{tab:metrics}, MAPLE outperforms UMAP across both measures. 
Both measures assume that the provided labels approximate ground truth semantics; consequently, higher scores in the low-dimensional space indicate better preservation of local decision boundaries and neighborhood relationships.

Neighborhood hit is a widely used measure in DR evaluation.
It quantifies the fraction of a point’s $k$NNs that share the same label, with higher values indicating that local neighborhoods are more semantically consistent. 
We set $k = 15$.
On the MNIST handwritten digits dataset~\cite{deng2012mnist}, where the classes are already well separated, MAPLE achieved a neighborhood hit~\cite{paulovich2008least} of 0.922 compared to 0.911 for UMAP.
This means that about 1\% more neighbors per point were correctly recovered as belonging to the same class. 
The improvement is 5\% more correct neighbors on the Fashion-MNIST dataset~\cite{xiao2017fashion}.
On STL-10, which has higher input dimensionality and greater intra-class variation than MNIST/Fashion-MNIST, MAPLE still improves neighborhood hit by roughly 4\% even with our minimal MLP encoder--projector. 
For single-cell data~\cite{packer2019lineage}, MAPLE achieved 0.554 vs. 0.450 on a \textit{Random} subset of \textit{C. elegans} cells, 0.626 vs. 0.561 on the \textit{Neuron} subset, and 0.750 vs. 0.718 on the \textit{Glia} subset.
Those improvements correspond to 5\%--10\% more correctly grouped neighbors in biologically meaningful subtypes when using MAPLE instead of UMAP.
This improvement is partly due to the high-resolution of the ground truth labels in this dataset.
Unlike broader categories, our evaluation uses 54 identified cell subtypes, which makes the measures more sensitive to MAPLE's ability to preserve fine-grained structure that other methods might blur.

$k$NN classification accuracy evaluates how well local neighborhoods preserve class identity and enables per-class analysis. 
For each data point, we predict its label via a weighted majority vote among its $k$ neighbors in the 2D layout ($k = 10$). 
This approach yields per-class accuracy scores defined as the mean prediction accuracy within each class (\cref{tab:combined_results}), while the overall score corresponds to the global mean accuracy across all data points (\cref{tab:metrics}).
MAPLE outperforms UMAP on overall per-dataset scores in \cref{tab:metrics}. 
To better understand these gains, we examine class-level performance in \cref{tab:combined_results}. 
On Fashion-MNIST, improvements primarily arise from resolving overlaps among visually similar classes such as `Shirt' and `Coat', with gains of 5--10\%. 
A similar trend appears in the \textit{C. elegans} neuron dataset, where MAPLE achieves a global improvement exceeding 11\%. 
For specific chemosensory lineages (e.g., ADF and ASE), MAPLE more clearly separates distinct cell subtypes (see~\cref{fig:case}). 
In contrast, for the mechanosensory OLQ lineage, 
improved separation of specific subtypes increases fine-grained accuracy while reducing scores for broader groupings.
This trade-off reflects MAPLE's tendency to resolve substructure rather than blur distinct cell types to preserve coarse lineage groupings.

\textbf{Other observations.} \quad 
Regarding global class separability, MAPLE generally produces layouts with clearer margins between classes compared to UMAP.
On global measures, both the DB~\cite{vergani2018soft} and CH~\cite{calinski1974dendrite} scores compare within-cluster scatter against between-cluster separation, and they favor compact, well-separated clusters.
MAPLE exhibits mixed performance compared to UMAP, depending on the data geometry. 
By revealing fine-grained classes, MAPLE often increases inter-class separation (thereby improving these scores); in some other cases, however, this increased resolution resulted in elongated structures that are penalized by these compactness-based measures.
MAPLE also achieves higher Hausdorff distances~\cite{huttenlocher2002comparing, taha2015metrics} across all experiments.

Clustering and perceptual measures can assess MAPLE's quality for downstream analytical workflows such as annotation and cluster analysis. 
For clustering measures, MAPLE outperforms UMAP on almost all datasets.
This confirms that the layouts produced by MAPLE are easier to partition and consistent with the improvement in local measures.
For perceptual measures (2AFC, ROC--AUC)~\cite{zhang2018unreasonable}, MAPLE and UMAP are comparable.
The two perceptual measures are ranking-based tests that ask whether same-class distances are consistently smaller than different-class distances, so they measure binary rankings of pairwise distances in the layout. 
Unless a method achieves substantially better class separation, layouts with rounded and compact clusters that UMAP often produces tend to score higher because their within-class distances are more uniformly small and stable. 
In contrast, MAPLE produces higher-resolution structures that can introduce greater variance within classes, which can lead to lower perceptual scores for datasets like Fashion-MNIST~\cite{xiao2017fashion}.

In summary, MAPLE generally outperforms UMAP across the selected quantitative measures. 
The magnitude of MAPLE's improvements depends on dataset complexity~\cite{isik2023information} and the granularity of the ground truth labels.
In this context, complexity refers to the difficulty of separating overlapping classes in high dimensions, while granularity refers to the level of detail in the annotations (e.g., broad categories vs. specific subtypes).
On well-separated datasets like MNIST handwritten digits, there is limited room for improvement; MAPLE shows modest quantitative improvements, whereas for single-cell data, the improvements are more pronounced.

\begin{figure}[t!]
  \centering
  \includegraphics[width=1\columnwidth]{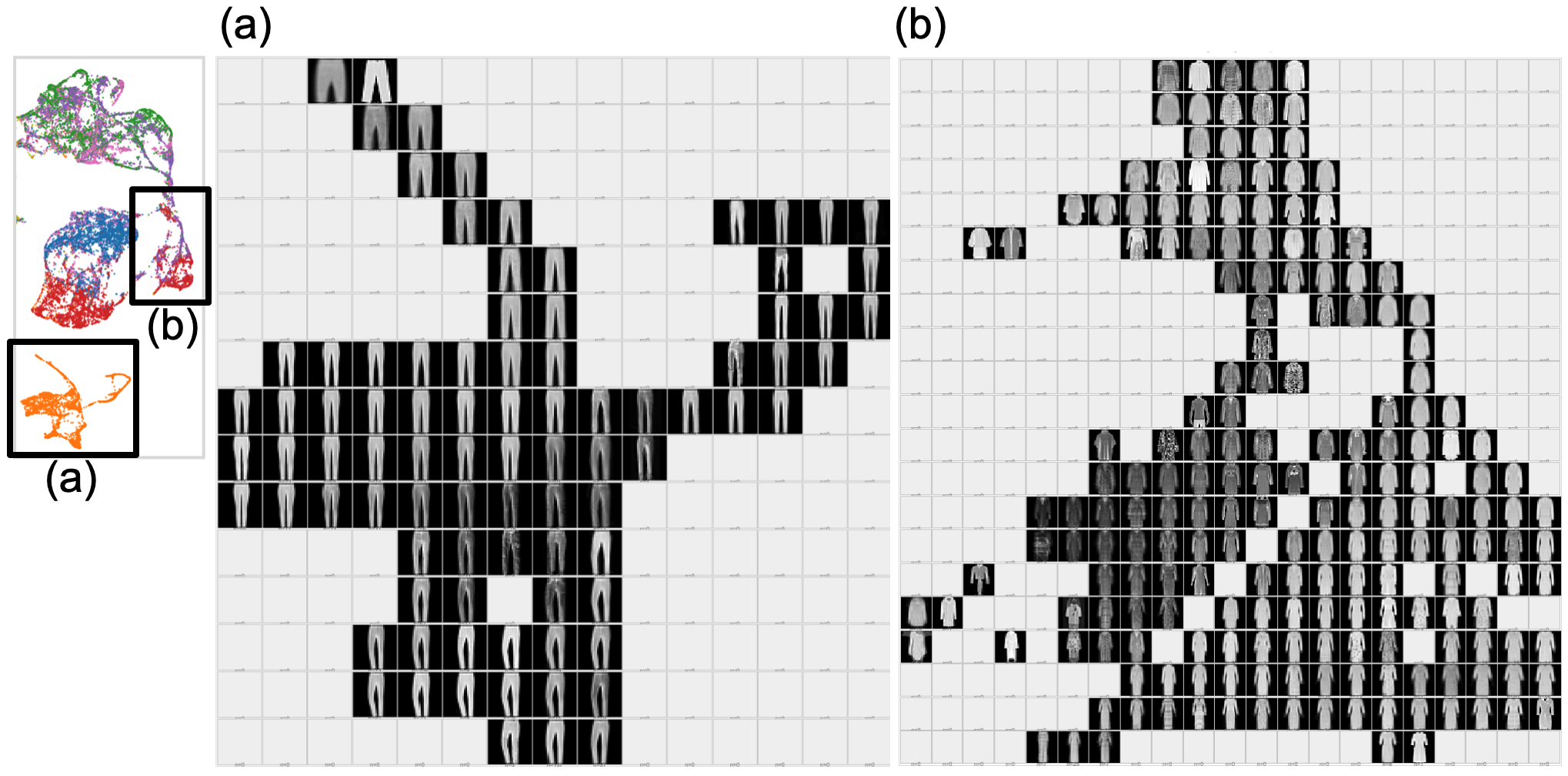}
  \caption{Closer visual inspection of MAPLE layouts on the Fashion-MNIST dataset (same layout as in \cref{fig:comp}). 
  (a) Underlying average images of the trouser class.
  (b) Underlying average images from the coat/dress region.
  }
  \label{fig:fashion-case}
\end{figure}

\subsection{Qualitative Evaluation}
We qualitatively compare the layouts produced by MAPLE with other DR methods. 
Since part of our pipeline is adapted from UMAP, we treat UMAP as the main baseline and provide several of its variants for comparison. 
In \cref{fig:comp}, each row represents one DR method, including MAPLE, UMAP~\cite{mcinnes_umap_2020} with Euclidean distance for input space graph construction, UMAP using cosine similarity, UMAP (Euclidean) with PCA~\cite{mackiewicz1993principal} preprocessing that reduces high-dimensional data to 50 principal components before applying UMAP, PCA, and $t$-SNE~\cite{maaten2008visualizing}. 

Across the datasets shown in \cref{fig:comp}, MAPLE produces layouts with noticeably finer internal structures. 
These finer internal structures contribute to clearer separation between broader categories, consistent with our quantitative results showing improved 2D neighborhood fidelity and boundary measures.
To examine whether these structures correspond to meaningful variation in the input data, we focus on the Fashion-MNIST dataset, as the underlying images require no additional domain knowledge to interpret. 
In \cref{fig:comp}, UMAP layouts under various configurations typically render this dataset as a set of largely uniform clusters where internal variations are difficult to identify. 
In contrast, MAPLE reveals a distinct geometric pattern across multiple runs.
For example, within the `Trouser' class (orange), MAPLE exposes two elongated structures in the upper region, a central compact group, and two smaller connected peaks at the bottom.

\textbf{Visual validation. \quad} 
To examine the patterns revealed by MAPLE and relate them to the underlying images, we designed a grid-based aggregation visualization (see \cref{fig:fashion-case}). 
For an interactively selected region of interest, we discretize the 2D space into a grid.
In each grid cell containing data points, we visualize the pixel-wise average of the corresponding images.
This analysis indicates that the structures produced by MAPLE correspond to visually meaningful fashion subtypes disentangled by the self-supervised objective. 
As shown in \cref{fig:fashion-case}(a), the upper-left branch in the `Trouser' class primarily contains wide-leg shorts, while the upper-right region captures thin, straight-leg cuts. 
The central cluster consists of standard jeans, and the bottom peaks correspond to trousers with one bent leg. 
Similarly, in the region containing the `Coat' and `Dress' classes (\cref{fig:fashion-case}(b)), MAPLE shows roughly three distinct subareas: coats at the top, dark clothing to the left, and skirts or dresses (identified by longer lower hems) at the bottom. 
Additional examples on the MNIST handwritten digits dataset are provided in the supplementary materials.

This ability to resolve intra-class variation is a key strength of MAPLE. 
By encouraging local neighborhoods to align with lower-rank structures through a self-supervised objective, MAPLE introduces perceptible gaps in the final layout between visually distinct sub-groups. 
From an information visualization perspective, these gaps align with the Gestalt law of proximity~\cite{7294427,bae2025uncovering}, supporting the perception of distinct groups and facilitating the identification of subtle data subtypes. 
While we do not claim that all geometric structures in the layout directly correspond to visually identifiable semantic components, the observed consistency between local layout structures and input-level visual patterns suggests that a substantial portion of the internal structures reflects coherent variation in the data and arises from the objective's suppression of redundant variance.
Overall, these results indicate that MAPLE enables higher-resolution visual exploration and supports the discovery of fine-grained intra-class patterns that are often obscured in other DR methods.

\subsection{Scalability and Runtime Analysis}

We evaluate the performance of MAPLE on a laptop equipped with an Apple M1 Pro chip and 16 GB of unified memory.
The training framework is implemented in PyTorch using the Metal Performance Shaders (MPS) backend to execute tensor operations on the integrated GPU.
To improve throughput, we use automatic mixed precision for gradient computations.
Due to current limitations in the MPS backend for certain singular-value decomposition operators, the nuclear norm required by the MMCR loss is computed on the CPU.
To accommodate large datasets under limited memory, we use a batch-loading strategy that avoids memory overflow.

\textbf{Data generation.} \quad 
To systematically evaluate scalability, we generate synthetic variations of the MNIST handwritten digits dataset~\cite{deng2012mnist} along two axes: feature dimensionality ($D$) and data size ($N$). 
To vary dimensionality, we spatially interpolate the original $28 \times 28$ pixel images to resolutions ranging from $8 \times 8$ up to $100 \times 100$, yielding feature vectors from $D=64$ to $D=10{,}000$.  
To assess scalability with respect to sample size, we expand the dataset from the original $10{,}000$ images up to $N=100{,}000$ via oversampling.
Each replicated sample is augmented using one of four basic transformations: Gaussian noise addition, rotation, random flips, or brightness adjustment.
All experiments use neighborhood sizes of $k=15$ and $k=30$, a batch size of $4,096$, and a training duration of $20$ epochs. 

\begin{figure}[t]
  \centering
  \includegraphics[width=1\columnwidth]{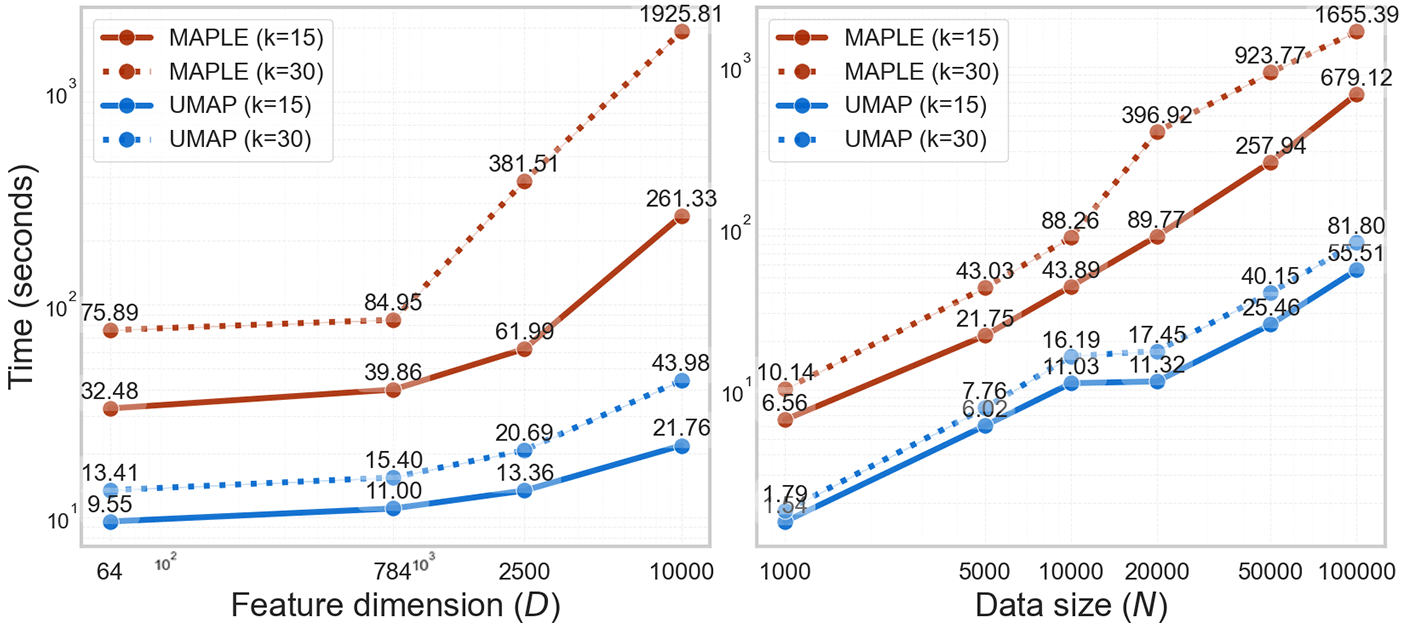}
  \caption{Completion time of MAPLE and UMAP in seconds. Left: runtime with varying feature dimensions ($D$) at fixed $N=10,000$. Right: runtime with varying data sizes ($N$) at fixed $D=784$. Solid and dotted lines indicate neighborhood sizes of $k=15$ and $k=30$, respectively. Both axes are shown on logarithmic scales, with tick labels $10^2$ and $10^3$ indicating orders of magnitude.}
  \label{fig:time}
\end{figure}

\textbf{Observation.} \quad 
The results are shown in \cref{fig:time}. 
MAPLE exhibits approximately linear runtime scaling with respect to both data size $N$ and feature dimensionality $D$, closely matching the scaling behavior of standard UMAP.
For instance, increasing the data size from $N=50{,}000$ to $N=100{,}000$ increases runtime by approximately 2.7 times (from roughly 4 minutes to 11 minutes). 
Runtime also scales linearly with $D$. 
However, we observe increased sensitivity to neighborhood size for high-dimensional inputs.
\textcolor{minor}{Beginning at $D=2{,}500$,} increasing $k$ from 15 to 30 leads to a significant runtime increase (\textcolor{minor}{e.g., }from 261 seconds to 1,925 seconds \textcolor{minor}{at $D=10{,}000$}).
This behavior arises from two factors.
First, the encoder processes entire local neighborhoods, causing the dominant computational cost to scale as $O(kND')$. 
Second, at high dimensionality, the resulting dense tensors exceed CPU cache capacity, making system memory bandwidth a limiting factor. 
While batching prevents memory overflow, the runtime bottleneck shifts to data loading and computation.
\textcolor{minor}{The bottleneck can be partially mitigated by reducing the batch size, although this introduces a trade-off with training efficiency~\cite{tsai2021selfsupervised}.}

The runtime difference between MAPLE and UMAP reflects an inherent trade-off in learning-based DR methods. 
MAPLE incurs additional computational cost due to the neural network training phase, resulting in a runtime of the order of minutes rather than seconds. 
However, this cost is predictable and scales linearly with dataset size.
In practice, MAPLE's runtime is comparable to other parametric DR methods~\cite{sainburg2021parametric}.
\textcolor{minor}{We advise practitioners to be mindful of feature dimensionality and neighborhood size when applying MAPLE, particularly in high-dimensional settings.}
Our results show that MAPLE can process datasets with up to $100{,}000$ samples on a standard laptop within minutes, without requiring high-performance hardware.

\begin{figure}[t!]
  \centering
  \includegraphics[width=1\columnwidth]{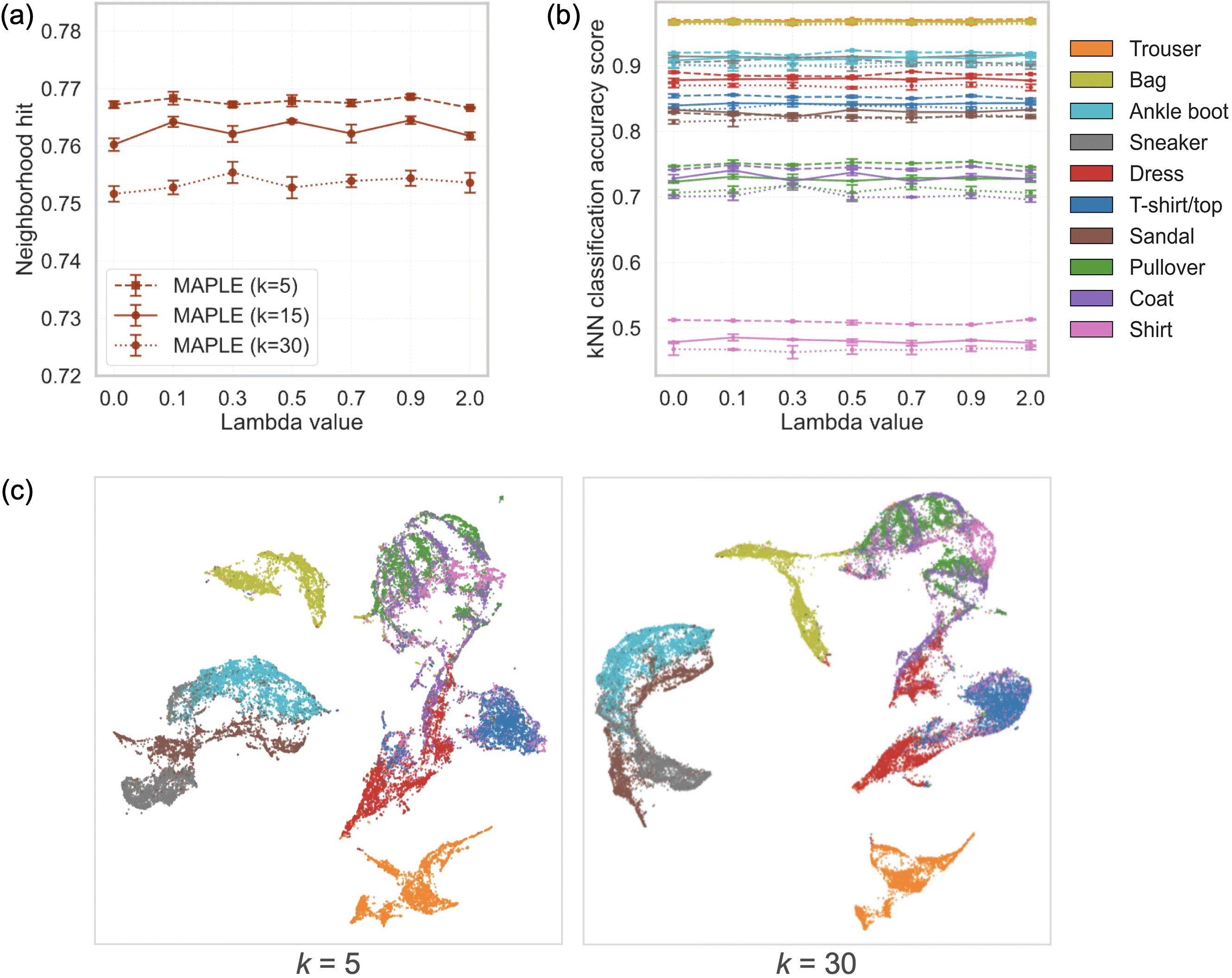}
  \caption{Parameter sensitivity analysis on the Fashion-MNIST dataset~\cite{xiao2017fashion} with $N = 40,000$. For (a) and (b), the x-axis denotes the parameter $\lambda$ and different line styles (e.g., dashed) denote neighborhood sizes $k \in \{5, 15, 30\}$, averaged across five independent runs. Vertical error bars for each data point represent the standard deviation across five runs.
  (a) Average neighborhood hit score~\cite{jeon2023zadu} (underlying data provided in the supplementary materials). 
  (b) Per-class $k$NN classification accuracy, where colors represent different class categories consistent with~\cref{fig:comp}.
  (c) MAPLE layouts for Fashion-MNIST with neighborhood sizes of $k = 5$ and $k = 30$.}
  \label{fig:ablation-result}
\end{figure}

\subsection{Parameter Sensitivity Analysis}  \label{sec:parameter} 
To assess the robustness of MAPLE with respect to its two primary hyperparameters, the neighborhood size ($k$) and the trade-off parameter ($\lambda$), we conducted a grid search analysis on the Fashion-MNIST dataset ($N=40{,}000$). 
The search space included $k \in \{5, 15, 30\}$ and $\lambda \in \{0, 0.1, 0.3, 0.5, 0.7, 0.9, 2.0\}$. We evaluated the resulting layouts using our two core quantitative measures, neighborhood hit and $k$NN classification accuracy, to examine whether MAPLE requires careful hyperparameter tuning or offers a broad, stable behavior.
The results are summarized in \cref{fig:ablation-result}, \textcolor{minor}{where different line styles (e.g., dashed) denote neighborhood sizes $k \in \{5, 15, 30\}$}.

\textbf{Observation.} \quad 
The results indicate that MAPLE remains stable across a wide range of hyperparameter settings (see~\cref{fig:ablation-result}). 
With a training duration of 20 epochs, all MAPLE configurations outperform the UMAP baselines across the evaluated measures (based on the full results underlying \cref{fig:ablation-result}, provided in the supplementary materials).
We observe that the trade-off parameter $\lambda$ has a relatively limited impact on the layout quality within the tested range (\cref{fig:ablation-result}(a) and (b)). 
This behavior is consistent with findings in the original MMCR study~\cite{yerxa2023mmcr}, where maximizing the global nuclear norm of centroids implicitly encourages compact local manifolds.
In our formulation, the global term provides strong separation, while the local term yields more modest refinements. 

In contrast, the neighborhood size $k$ plays a more influential role. 
MAPLE achieves its highest quantitative scores at smaller neighborhood sizes ($k=5$), as neighborhood hit is a highly local metric.
This behavior is also reflected qualitatively in \cref{fig:ablation-result}(c), where structurally similar classes such as `\textcolor{minor}{Ankle} Boot', `Sandal', and `Sneaker' (the blue--brown--grey cluster on the left) appear more clearly separated than at $k=30$. 
We hypothesize that smaller $k$ values yield smaller and more homogeneous neighborhood ``views'', which are better aligned with the locality assumptions in MMCR-style self-supervised learning~\cite{yerxa2023mmcr, isik2023information}.
To contextualize these results, we further compare MAPLE against UMAP using matched neighborhood sizes; detailed neighborhood hit scores for each method are reported in the supplementary materials.
MAPLE provides neighborhood improvements at smaller $k$, while larger $k$ values yield stronger relative gains.
During development, we also explored soft probabilistic $k$NN assignments, but observed only marginal improvements over the standard formulation.
This suggests that binary selection of high-confidence neighbors is sufficient for MAPLE to learn meaningful manifold structure, offering a favorable balance between simplicity and accuracy.

\section{Applying MAPLE to Single-cell Datasets}
\label{sec:cases}
\begin{figure}[t!]
  \centering
  \includegraphics[width=1\columnwidth]{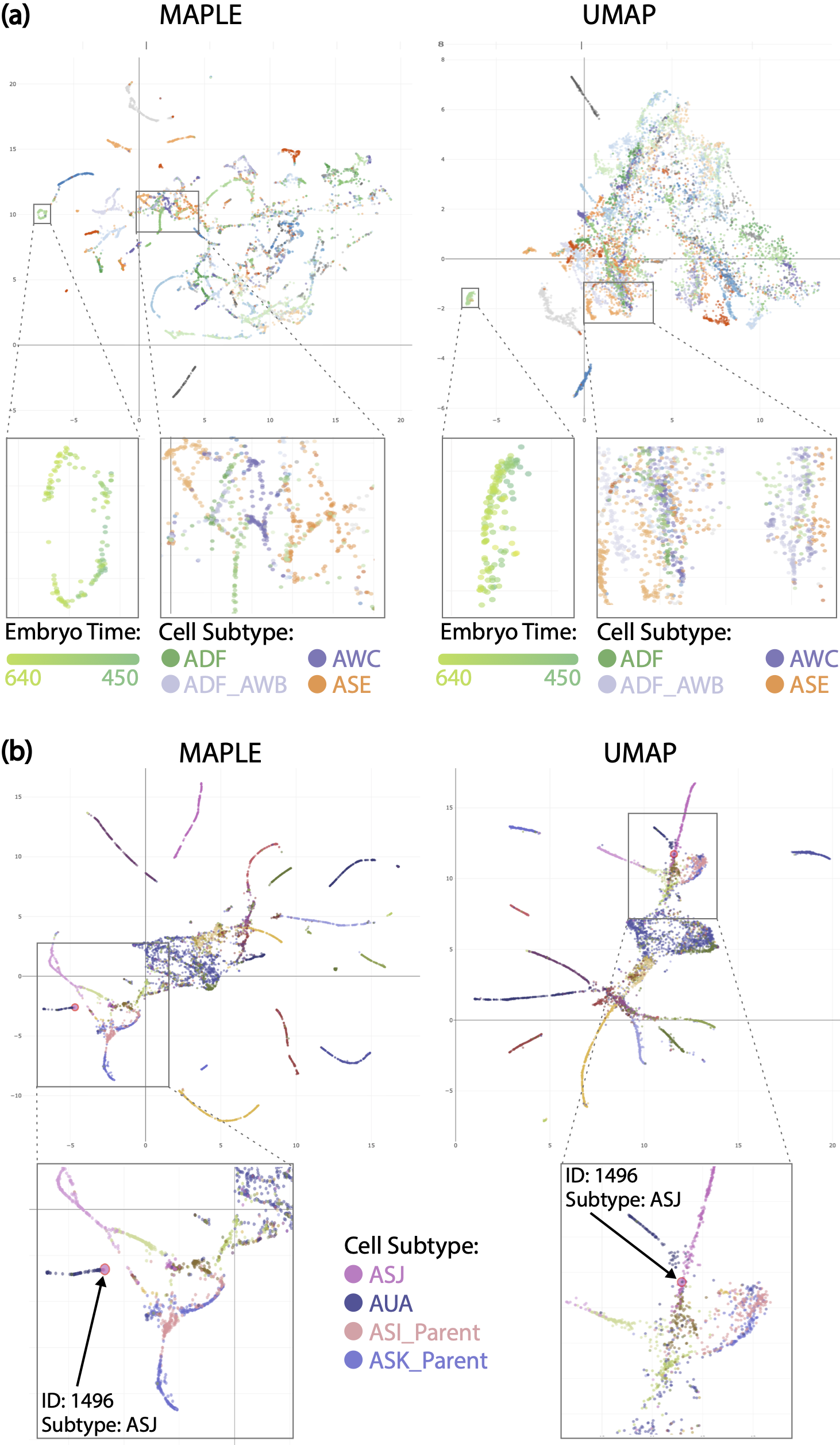}
  \caption{Comparison of MAPLE and UMAP layouts of \textit{C. elegans} neuron data.
(a) Raw neuron subset~\cite{packer2019lineage} (20,222 gene dimensions; corresponds to \cref{fig:comp}, column 5). MAPLE shows clearer lineage-aligned separability than UMAP, consistent with embryo time and cell subtypes.
(b) Monocle-preprocessed neuron subset (50 PCA dimensions; corresponds to \cref{fig:comp}, column 7). MAPLE provides cell subtype disambiguation and reveals transitional ``bridge'' cells that are obscured in UMAP.}
  \label{fig:case}
\end{figure}

As DR methods play a central role in single-cell bioinformatics~\cite{becht2019dimensionality,narayan_assessing_2021,moon_PHATE_2019,kobak_initialization_2021,chari2023specious}, we demonstrate how MAPLE supports expert-driven exploratory analysis in this domain.
We present an expert-guided evaluation, in which a biologist inspected MAPLE and UMAP layouts side by side using our interactive visual tool.
The tool provides standard exploration functionalities, such as zooming and panning, brushing and linking, and details-on-demand.
Through this process, the expert confirmed that MAPLE supports more reliable bioinformatics analyses by validating existing annotations and aligning with established single-cell knowledge. 

\textbf{Data exploration in \textit{C. elegans} embryogenesis.} \quad
We selected the \textit{C. elegans} dataset because every individual worm develops through the same sequence of cell divisions, and its cell lineage has been largely mapped. 
These characteristics make it ideal for closely examining the DR layouts to see how they can help describe gene expression patterns and trace transcriptome dynamics during differentiation and development. 
The dataset we use is from Packer et al.~\cite{packer2019lineage}, which contains 86,204 cells with 20,222 measured gene expression dimensions, spanning a wide diversity of cell types.
It is fully annotated, and many subsequent studies~\cite{fung2020cell, sivaramakrishnan2023transcript} have validated the accuracy of these annotations.
For each cell, annotations are provided at both broad levels (e.g., neurons and glia) and finer subtypes (e.g., amphid sensory neurons such as ADF, AWB), 
together with lineage information such as ``parent'' or ``neuroblast'' states, and rich metadata for analysis. 
These annotations facilitate comparison of proximity and expression differences across multiple scales of analysis. 

In our study, we focused on two major subsets: neurons and glia.
The expert interactively explored the layouts produced by MAPLE and UMAP using Euclidean metrics. 
\cref{fig:case}(a) shows the MAPLE layout of the raw neuron and glia subsets with full metadata (displayed upon mouse hover or box selection), where the neuron population is further divided into more than 20 fine-grained cell subtypes, and the top panels are colored accordingly.
\cref{fig:case}(b) shows another set of \textit{C. elegans} data after standard preprocessing with Monocle~\cite{trapnell2014dynamics} (applying PCA~\cite{mackiewicz1993principal}, batch correction~\cite{haghverdi2018batch}, and regression of the continuous background contamination factors), which reduced the dimensionality from 20,222 to 50 and the granularity of annotations into a smaller set of developmental categories. 
The side-by-side comparison allowed us to examine how MAPLE preserves detailed lineage information in both raw and preprocessed data while still producing meaningful substructures.

\textbf{MAPLE brings better cell subtype disambiguation.} \quad 
During the evaluation, the expert confirmed that the separations produced by MAPLE align mostly with the known \textit{C. elegans} cell lineage. 
For example, as shown in \cref{fig:case}(a), MAPLE provides clearer separation of both embryo time and known cell subtypes compared to UMAP. 
Amphid sensory neurons such as ADF and AWC, which often appear clustered together in UMAP, are also resolved into distinct trajectories in MAPLE. 
At the same time, MAPLE preserves alignments between closely related categories such as ADF and their progenitor ADF\_AWB, which are consistent with developmental biology findings~\cite{packer2019lineage}. 
This combination of proper separation and alignment with biological events gives domain experts confidence that MAPLE's higher-resolution results more faithfully capture lineage structure than UMAP.

\textbf{MAPLE facilitates hypothesis generation.} \quad 
Beyond validation, MAPLE also supports the discovery of potential new insights.
As illustrated in \cref{fig:case}(b), MAPLE not only disambiguates progenitor states such as ASI\_parent and ASK\_parent, but also highlights individual cells that act as ``bridges'' between trajectories~\cite{packer2019lineage}.
For instance, a cell that UMAP merges into the AUA cluster is instead positioned by MAPLE between ASJ and AUA lineages. 
Such bridge-like cells may represent transitional populations where progenitors begin differentiating, which may be captured through the expression of marker genes of both descendant types~\cite{wagner_lineage_2020}.
Identifying these rare states is valuable because it enables researchers to generate and test hypotheses about the molecular signatures of ancestor cells and the order in which lineages branch. 
Those insights would remain hidden in other DR layouts.

MAPLE can be used as a stand-alone method, or as a supplementary panel to mitigate the ``hallucination'' of connectivity that has been reported for UMAP~\cite{sun2019accuracy}. 
While UMAP is widely used in single-cell studies, it has also been criticized for generating overly confident continuous trajectories that may not exist~\cite{sun2019accuracy, xia2024statistical,chari2023specious}. 
This can mislead analysts into forming incorrect hypotheses, especially in datasets where annotations are lacking and uncertainty is high. 
MAPLE's local compression and diversification strategy produces clearer separations, which may raise new questions about connectivity while also providing additional perspective to help validate or challenge the results of other DR methods.

\section{Discussion and Future Work}
\label{sec:discussion}
While UMAP remains highly competitive in terms of efficiency and overall robustness, MAPLE offers a new perspective by explicitly learning fuzzy neighborhood relations through self-supervised learning.

Conceptually, MAPLE belongs to a growing line of work that focuses on the often-overlooked yet critical aspect of input-space modeling in DR methods~\cite{bohm2022unsupervised,wang2025dimension}.
By reinterpreting neighborhood graphs as implicit views, MAPLE establishes a connection between MVSSL and DR, while introducing geometry-aware objectives to regularize local structure. 
At the same time, MAPLE represents a hybrid approach, uniquely combining parametric graph construction with non-parametric projection.
We believe this opens up broader discussions on the multifaceted relationships between DR and SSL, as discussed in prior theoretical and empirical studies~\cite{hu2023your, damrich2022tsne}.

Furthermore, MAPLE highlights the need to reconsider how DR methods are evaluated. 
For methods like ours, quantitative evaluation should move beyond statistical consistency checks between input and output space and instead develop new measures---potentially inspired by computer vision and human perception~\cite{taha2015metrics, zhang2018unreasonable, bae2025uncovering}---that can better capture downstream, task-oriented goals. 
In addition, approaches developed in medical image segmentation~\cite{taha2015metrics}, where evaluation is designed to quantitatively reflect perceptual quality and task performance, could potentially be adapted to expand the current set of evaluation measures in MAPLE.  

\textbf{Strengths and limitations across data characteristics.} \quad
A key characteristic that differentiates MAPLE from existing DR methods~\cite{espadoto2019toward} is its explicit focus on resolving intra-cluster variation. 
By revealing higher-resolution structures within clusters, MAPLE makes internal organization more visible in the resulting layouts, which often also increases overall class separability, as shown in our quantitative evaluation results. 

If the primary objective is class separability, beyond the experiments performed on real-world datasets presented in \cref{tab:metrics}, we observe that MAPLE's effectiveness depends strongly on the structure of the input data.
We define ``simple'' datasets as those with already high manifold capacity~\cite{yerxa2023mmcr} and clean geometry (e.g., synthetic manifolds like the Helix or Swiss roll).
On such simple datasets, MAPLE tends to provide limited additional benefit, as a standard $k$NN graph already captures the local manifold structure reasonably well. 
In contrast, MAPLE consistently outperforms standard methods on more ``complex'' datasets (e.g., MNIST, Fashion-MNIST), where neighborhoods exhibit noise and significant class overlap. 
These results suggest that MAPLE is especially effective when the graph construction benefits from additional geometric regularization.

Furthermore, we observe that preprocessing and the resulting dimensionality influenced MAPLE's performance.
Theoretical and empirical results on manifold capacity~\cite{schaeffer_towards_2024}, together with our MAPLE experiments, suggest that performance degrades when data are heavily whitened and the embedding dimension is large but contains little orthogonal variance.
For instance, after applying PCA and retaining more than 200 principal components on a simple dataset, if neighborhoods are fuzzy, MAPLE may degenerate, meaning that the global nuclear norm has no additional variance to distribute, so centroids in the embedding fail to spread out, and the low-dimensional layout contracts into compact clusters.

In summary, MAPLE is more effective for data where local neighborhoods exhibit rich, high-dimensional variation within clusters, and less effective when data are heavily whitened or already well modeled by linear similarities. 
Exploring these boundaries systematically and connecting them to manifold capacity theory remains an interesting direction.
Meanwhile, multimodal datasets (e.g., image--text), where heterogeneous manifolds are common, present a promising application area for future work with MAPLE.

\textbf{Broader connections.} \quad
From an information-theoretic perspective, MAPLE conceptually achieves lower information loss about neighborhood connectivity compared to UMAP. 
Although one of MAPLE's objectives is to compress neighborhoods, what is compressed are noisy or curved directions that do not contribute to the intrinsic manifold geometry. 
At the same time, MAPLE enhances global separability by suppressing irrelevant local variance and reallocating it toward informative between-neighborhood differences.
In this way, MAPLE preserves more task-relevant information, and the resulting low-dimensional layouts reveal a finer intra-cluster structure. 
This perspective also suggests opportunities for future work on developing information-theoretic measures that evaluate not only the projection process, but also the role of input-space modeling and graph refinement in DR. 
Developing entropy- or mutual-information-based measures could help quantify how much structure is maintained or lost at each stage, and potentially offer a principled way~\cite{galvez2023role} to compare methods that differ in their treatment of neighborhood graphs or general input-space modeling. 

Moreover, since MAPLE provides a parametric geometry-aware approach to construct graphs and is fully compatible with UMAP's CE optimization, the same pipeline could in principle be adapted to other graph-based DR methods. 
Since methods such as $t$-SNE~\cite{maaten2008visualizing} and Laplacian Eigenmaps~\cite{belkin2003laplacian} begin with a neighborhood graph or similarity matrix before applying their respective optimization objectives, MAPLE could serve as a modular graph refinement stage that improves the reliability of the neighborhood structure while leaving the downstream projection objectives unchanged. 
This opens the door to integrating geometry-aware SSL objectives into a broader range of graph-based DR methods. 

Finally, through our use cases, we see potential for further research on using geometry-aware SSL to improve DR in practice. 
In particular, such approaches can produce layouts with finer internal resolution, supporting closer inspection of subtle structure in complex, high-dimensional data.
In biological applications, most existing workflows rely on linear preprocessing steps before applying DR~\cite{haghverdi2018batch, sun2019accuracy}. 
Our findings suggest that MAPLE brings the most quantitative improvements when applied directly to raw single-cell data, indicating that SSL-based DR methods could be further developed and adapted to domain-specific settings. 
More broadly, MAPLE illustrates how geometry- or manifold-aware SSL can enhance the expressiveness of DR and contribute to uncovering new insights in scientific data.

\section{Conclusion}
\label{sec:conclusion}
In this work, we introduced MAPLE, a new nonlinear dimensionality reduction method that extends UMAP with self-supervised learning objectives for neighborhood graph modeling. 
Through the reinterpretation of maximum manifold capacity representations, MAPLE effectively improves neighborhood construction in the embedding space. 
Our quantitative and qualitative evaluations show that MAPLE leads to clearer visual cluster separations and finer resolution of subclusters compared to UMAP. 
These results suggest that MAPLE provides a promising direction for geometry-aware SSL in DR for visual analysis. 
Future work would include addressing scalability through parallelization and further optimization strategies, providing formal theoretical explanations, developing new task-oriented DR measures, and extending the framework to broader graph-based DR methods as well as multimodal contexts.

\section*{Acknowledgments}
The authors wish to thank the anonymous reviewers for their valuable and insightful feedback.
This work was supported through the ELLIIT environment for strategic research in Sweden.

\bibliographystyle{abbrv-doi-hyperref}

\bibliography{maple}


\end{document}